\documentclass[10pt,twocolumn,letterpaper]{article}

\usepackage[pagenumbers]{cvpr} 

\definecolor{cvprblue}{rgb}{0.21,0.49,0.74}
\usepackage[pagebackref,breaklinks,colorlinks,allcolors=cvprblue]{hyperref}

\newcommand{\ourmodel}{SaPaVe}
\newcommand{\ourdataset}{ActiveViewPose-200K}
\newcommand{\oursimbenchmark}{ActiveManip-Bench}

\usepackage{wrapfig}
\usepackage{pifont}
\usepackage{graphicx}
\usepackage{tabularx}
\usepackage{multirow}
\usepackage{booktabs}
\usepackage{amsmath}  
\usepackage{amssymb}  
\usepackage{mathrsfs} 
\usepackage{amsthm,amsfonts}
\usepackage[table,dvipsnames]{xcolor}
\usepackage{array}
\usepackage{float}
\usepackage{pifont} 
\usepackage{bbding} 
\usepackage{fontawesome} 
\usepackage[most]{tcolorbox}
\usepackage{listings}
\usepackage{etoolbox}

\def\ie{\emph{i.e.}}
\def\eg{\emph{e.g.}}

\definecolor{ourpurple}{HTML}{7030A0} 
\definecolor{ouryellow}{HTML}{F4B402} 

\definecolor{codeblue}{rgb}{0.25,0.5,0.5}
\definecolor{mylightblue}{rgb}{0.96, 0.995, 1}
\definecolor{myblue}{rgb}{0.88,0.98,1}
\definecolor{mygreen}{rgb}{0.92, 1.0, 0.92}
\definecolor{myred}{rgb}{1, 0.9, 0.9}
\definecolor{mygray}{gray}{0.95}
\definecolor{mydarkblue}{rgb}{0,0.08,1}
\definecolor{mydarkred}{rgb}{0.8,0.02,0.02}
\definecolor{mydarkorange}{rgb}{0.40,0.2,0.02}
\definecolor{mypurple}{RGB}{239,229,253}
\definecolor{mygold}{rgb}{0.75,0.6,0.12}
\definecolor{mydarkgray}{rgb}{0.66, 0.66, 0.66}
\definecolor{mydarkgreen}{rgb}{0.02,0.6,0.02}
\definecolor{mygray}{gray}{0.9}
\definecolor{tablegray}{gray}{0.95}

\definecolor{codegreen}{rgb}{0,0.6,0}
\definecolor{codegray}{rgb}{0.5,0.5,0.5}
\definecolor{codepurple}{rgb}{0.58,0,0.82}
\definecolor{backcolour}{rgb}{0.95,0.95,0.92}

\lstdefinestyle{myprompt}{
    backgroundcolor=\color{backcolour},   
    commentstyle=\color{codegreen},
    keywordstyle=\color{magenta},
    numberstyle=\tiny\color{codegray},
    stringstyle=\color{codepurple},
    basicstyle=\ttfamily\footnotesize,
    breakatwhitespace=false,         
    breaklines=true,                 
    captionpos=b,                    
    keepspaces=true,                 
    numbers=left,                    
    numbersep=5pt,                  
    showspaces=false,                
    showstringspaces=false,
    showtabs=false,                  
    tabsize=2,
    frame=single
}

\def\paperID{1271} 
\def\confName{CVPR}
\def\confYear{2026}

\title{SaPaVe: Towards Active Perception and Manipulation \\ in Vision-Language-Action Models for Robotics}

\author{
Mengzhen Liu\textsuperscript{1,3}\footnotemark[1]~, \quad
Enshen Zhou\textsuperscript{2,3}\footnotemark[1]~~\footnotemark[3]~, \quad
Cheng Chi\textsuperscript{3}, \;
Yi Han\textsuperscript{2,3}, \; \\
Shanyu Rong\textsuperscript{1,3}, \; 
Liming Chen\textsuperscript{3}, \; 
Pengwei Wang\textsuperscript{3}, \;
Zhongyuan Wang\textsuperscript{3}, \;
Shanghang Zhang\textsuperscript{1,3}\footnotemark[2]\\
\small $^{1}$State Key Laboratory of Multimedia Information Processing, School of Computer Science, Peking University~~ \\
\small$^{2}$School of Software, Beihang University ~~ \small$^{3}$Beijing Academy of Artificial Intelligence ~~\\
\tt\footnotesize mengzhenliu25@stu.pku.edu.cn~~~zhouenshen@buaa.edu.cn~~~shanghang@pku.edu.cn\\
}

\begin{document}
\twocolumn[{%
\renewcommand\twocolumn[1][]{#1}%
\maketitle
\vspace{-10mm}
\begin{center}
    \captionsetup{type=figure}
    \includegraphics[width=1\linewidth]{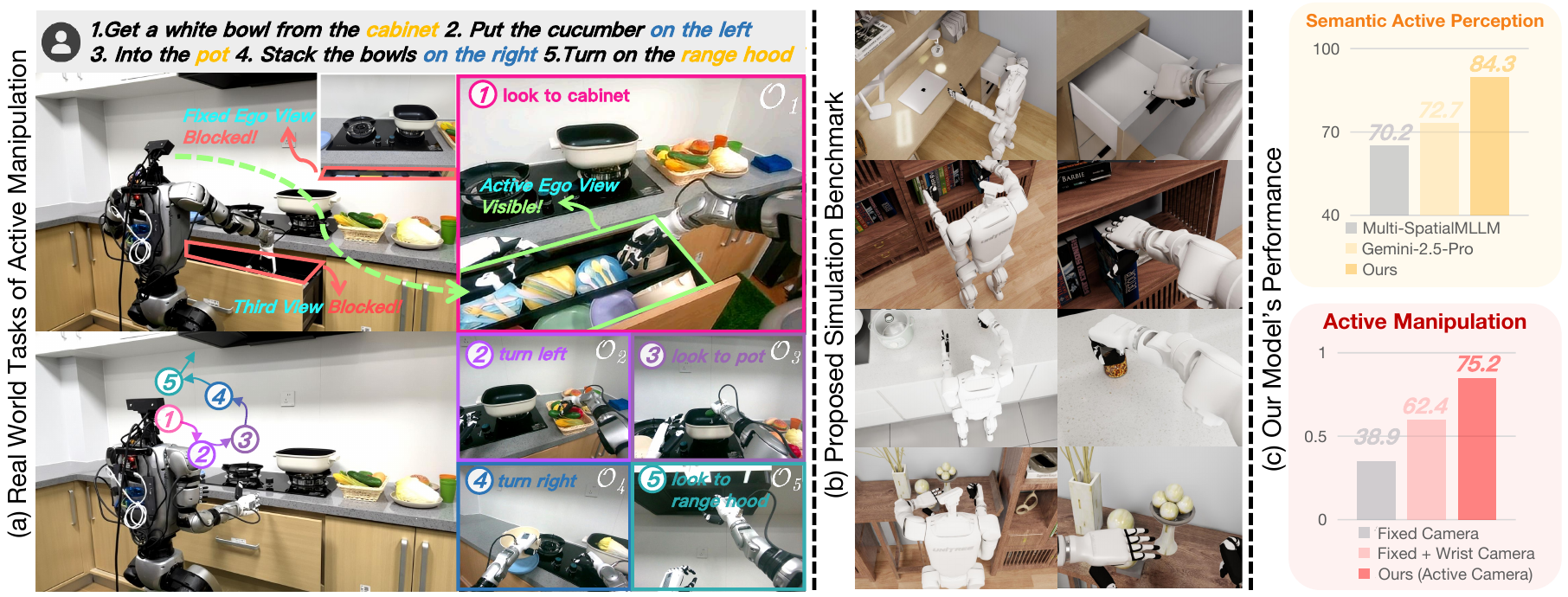}
    \vspace{-5mm}
    \captionof{figure}{    
    We propose {\ourmodel}, an end-to-end active manipulation framework that jointly integrates semantic active perception and active-view execution; the former selectively shifting viewpoints to reveal task-critical cues in cluttered scenes, while the latter grounds newly acquired observations into immediate actions, enabling success even from suboptimal views.
    \textbf{(a)} For instance, grasping the white bowl in \(\mathcal{O}_1\) requires rotating the egocentric view, as both fixed ego view and third-person view are occluded. In contrast, targeting the range hood handle (\(\mathcal{O}_5\)) only needs a brief upward shift, since precise centering is unnecessary and awkward to reach. 
    \textbf{(b)} To address the limitations of fixed-view benchmarks and the cost of real-world trials, we introduce {\oursimbenchmark}, a richly annotated benchmark spanning 12 tasks, 100 objects, and 20 diverse scenes.
    \textbf{(c)} On this benchmark, {\ourmodel} outperforms all baselines with an average success rate of 75.2\%.
    }\label{fig: motivation}
\end{center}%
}]

\let\thefootnote\relax\footnotetext{$^*$ Equal contribution\hspace{3pt} \hspace{5pt}$^\dagger$ Corresponding author\hspace{5pt} $^\ddagger$ Project leader
}

\begin{abstract}

Active perception and manipulation are crucial for robots to interact with complex scenes.
Existing methods struggle to unify semantic-driven perception actively with robust, viewpoint-invariant execution accordingly.
To this end, we propose \ourmodel, an end-to-end framework that jointly learns these capabilities in a data-efficient manner.
Central to our approach is a decoupling of camera and manipulation actions, contrary to shared-action-space, and learning in a bottom-up strategy: we first train semantic camera control on our proposed large-scale dataset, then jointly optimizes both action types via hybrid data.
To support this, we introduce \ourdataset, comprising 200k image-language-camera movement pairs for semantic camera movement learning, and a 3D geometry-aware module that improves execution robustness under dynamic viewpoints.
We further present \oursimbenchmark, the first benchmark filling the gap to evaluate active manipulation.
Extensive experiments in both simulation and real-world settings show that \ourmodel\ outperforms recent VLA models such as GR00T N1 and $\pi_0$, achieving up to 31.25\% higher success rates in real-world tasks.
Our results show that tightly coupled perception and execution, when trained with decoupled yet coordinated strategies, enable efficient and generalizable active manipulation.
Please see the project page at \href{https://lmzpai.github.io/SaPaVe}{https://lmzpai.github.io/SaPaVe}.

\end{abstract}    
\vspace{-3mm}
\section{Introduction}

It is a long-lasting goal to create embodied agents that can actively perceive complex, dynamic scenes and act accordingly in a human-like manner~\cite{li2025amo,ViA}.
%
Achieving such active manipulation is challenging, as it demands two distinct yet complementary abilities, \textit{semantic active perception} and \textit{active-view execution}, as shown in \cref{fig: motivation}.
The former involves strategically adjusting viewpoints to acquire relevant information in cluttered scenes (\eg, shifting the ego-view to reveal a bowl hidden inside a cabinet, reducing occlusions).
The latter grounds newly obtained perceptual cues into immediate actions, ensuring successful task completion, even from suboptimal viewpoints (\eg, briefly glancing upward to locate a range hood handle without centering it is sufficient to initiate interaction).
Crucially, effective active manipulation hinges on the tight coupling of these two abilities, enabling the agent to perceive purposefully and act meaningfully, with both informing each other in real time.

Recent advances in Vision-Language Models (VLMs) \cite{yang2025qwen3,comanici2025gemini,liu2024deepseek,zhang2025beyond} have improved semantic instruction understanding.
A common approach leverages VLMs for \textit{active perception} by formulating it as a Visual Question Answering (VQA) task, selecting the best viewpoint from a discrete set of candidates~\cite{wang2025roboretriever,sripada2024ap}.
However, this discretization hinders fine-grained camera control and manipulation, as it fails to connect high-level semantics with the continuous camera pose space.
End-to-end Vision-Language-Action (VLA) models~\cite{gr00t,pi0} aim to bridge this gap, but are typically trained on fixed near-optimal head-camera views, making them sensitive to viewpoint shifts and limiting their robustness in \textit{active-view execution}.
Moreover, collecting additional data with both head-camera and action labels in the real world \cite{activeumi,open-television,lu2025mobile,ze2025twist2} is expensive and constrained, so simply fine-tuning VLA models in a unified action space leads to conflicts and suboptimal performance.

In this work, we introduce \textbf{\ourmodel} (\textbf{S}emantic \textbf{a}ctive \textbf{P}erception and \textbf{a}ctive-\textbf{V}iew \textbf{e}xecution), an end-to-end active manipulation framework that can achieve semantic active perception and active-view execution simultaneously, not only benefiting from the semantic power of VLMs, but also enjoying control over both head camera and manipulation actions in a data-efficient manner.
Our key insight is that camera movement is embodiment-agnostic and easier to learn, and we design a bottom-up learning strategy: we first learn camera movement independently using a large-scale dataset, then jointly optimize camera movement and other action generation via a decoupled action space using a mix of manipulation datasets.
This decoupling minimizes cross-interference and allows each action type to benefit from specific data.
%

To be specific, we first propose \textbf{{\ourdataset}}, a high-quality dataset comprising 200k image-language and camera movements pairs, enriched with highly detailed semantic annotations to enable task-oriented camera movement learning.
In the first stage, we use this dataset to supervise only camera movement and introduce a camera adapter to retain semantic camera movement knowledge to support \textit{active perception}.
Meanwhile, we design a simple yet effective universal spacial knowledge injection approach to 3D geometry information, supporting \textit{active-view execution} learning in the second stage, even under dynamic viewpoint changes induced by active perception.
In this second stage, we leverage diverse hybrid data to achieve robust, semantically grounded active manipulation, guided by the pretrained camera adapter, with high data efficiency.

To comprehensively evaluate {\ourmodel}, we conduct extensive experiments in both simulation and real-world settings.
First, we assess our semantic active perception capability on the {\ourdataset} and surpass Gemini 2.5 Pro~\cite{comanici2025gemini} by 16\% despite having only 2B parameters.
To address the limitations of fixed-viewpoint manipulation evaluation, we introduce the first simulated active manipulation benchmark, featuring 12 richly annotated tasks across 100 objects and 20 diverse scenes.
On this benchmark, our model achieves the best performance, even surpassing fixed-view VLA (\eg, GR00T-N1~\cite{gr00t}) by an absolute 58\% success rate.
Moreover, we collect real-world active manipulation data and directly fine-tune existing VLA models (e.g., $\pi_0$~\cite{pi0}, GR00T-N1~\cite{gr00t}) with a unified action space.
Our model still outperforms these baselines, exceeding $\pi_0$ by 40\% and GR00T-N1 by 31.25\%.
In summary, our contributions are threefold:
\begin{itemize}
    \item We propose \textbf{\ourmodel}, a novel end-to-end framework that first achieves active manipulation with a bottom-up learning strategy in a data-efficient way.

    \item We introduce \textbf{\ourdataset}, a well-annotated dataset tailored for learning semantic camera control, and \textbf{\oursimbenchmark}, the first benchmark that fills the gap in evaluating active manipulation.

     \item Extensive experiments show that \textbf{\ourmodel} excels in active perception and showcases impressive active manipulation capability in simulation and real-world settings.
\end{itemize}
\section{Related Work}

\noindent\textbf{Active Perception and Manipulation.}
Active perception and manipulation~\cite{bajcsy1988active,ViA} is a long-standing research area in robotics and computer vision.
Traditionally, this problem is formulated as a Next-Best-View (NBV) task~\cite{breyer2022closed,zhang2023affordance,krainin2011autonomous,connolly1985determination,isler2016information,bircher2016receding,naazare2022online}, where an agent iteratively selects a viewpoint to maximize an information gain metric. 
However, these methods often rely on non-end-to-end pipelines and lack semantic input, rendering them computationally prohibitive and incapable of adapting to diverse, semantically rich tasks.
With the advent of VLMs~\cite{yang2025qwen3,zhou2025robotracer,zhou2025roborefer,team2025robobrain,han2025tiger,zhou2024code,zhou2024minedreamer,qin2024mp5,comanici2025gemini,tan2026robobrain,achiam2023gpt}, recent works have explored new paradigms. 
Some regard active perception as a VQA task~\cite{wang2025roboretriever,sripada2024ap}, where a VLM evaluates a discrete set of candidate camera motions to select the optimal one. 
A significant limitation of this approach is its inability to operate within a continuous camera motion space, constraining the agent's flexibility. 
To address this, the rapid development of VLA models~\cite{pi0,gr00t,kim2025openvla} has spurred interest in end-to-end solutions that jointly model camera control and manipulation actions. 
Nevertheless, directly augmenting the action space of a VLA to include continuous camera movement~\cite{activeumi,yu2025egomi} exacerbates the demand for large-scale training data. 
Such demonstration data, featuring concurrent camera movement and manipulation, is exceptionally scarce, hindering the model's ability to acquire a genuine active manipulation capability. 
In contrast, our work proposes a bottom-up, end-to-end strategy that endows policies with active manipulation skills in a data-efficient manner.

\vspace{+1mm}
\noindent\textbf{VLA for Robotics.}
VLA has made significant strides in recent years. 
However, The majority existing VLA models~\cite{gr00t,pi0,liu2024robomamba,liu2025hybridvla,liu2024robomamba2}, are trained and evaluated under the assumption of a fixed camera viewpoint that is near-optimal.
This setting largely overlooks the critical necessity of active perception for handling visual ambiguity and occlusions in real-world scenarios.
While some concurrent works~\cite{activeumi,yu2025egomi} attempt to naively finetune VLA for both camera and other control, they face two primary challenges.
\textbf{(1)} Data Scarcity: Existing VLA training datasets~\cite{wu2024robomind,o2024open,bu2025agibot} mainly consist of demonstrations from a fixed perspective. 
The high cost and complexity of collecting active perception and manipulation data in the real world make it difficult to scale.
\textbf{(2)} Lack of 3D Geometric Awareness: Current VLA models~\cite{bu2025univla,rt1} often fail to fully leverage 3D geometric priors. This makes them highly sensitive to camera perturbations and unable to reason effectively under viewpoint changes.
To address these challenges, we propose a data-efficient training strategy that systematically incorporates diverse 3D geometric priors, which augment active manipulation capabilities and fortify the model's robustness against viewpoint variations.

\vspace{+1mm}
\noindent\textbf{Benchmarks for Active Perception and Manipulation.}
Existing active perception and manipulation studies are primarily evaluated in real-world using a single robot arm with a wrist-mounted camera~\cite{wang2025roboretriever,breyer2022closed,zhang2023affordance,sripada2024ap,liu2024segment,yuan2025motiontrans,fu2025cordvip}, a humanoid robot equipped with a customized 2-DoF head~\cite{activeumi,yu2025egomi,ze2025twist2,open-television,li2025amo}, or a humanoid robot with a modified 6-DoF head~\cite{ViA}.
However, such evaluations are difficult to reproduce due to the inherent challenges in replicating identical real-world experimental setups.
In contrast, simulation environments~\cite{isaac,todorov2012mujoco,xiang2020sapien} offer perfect reproducibility and standardized testbeds. 
However, the vast majority of existing simulation-based benchmarks~\cite{liu2023libero,mees2022calvin} are restricted to fixed camera viewpoints. 
Consequently, there is currently a noticeable absence of established benchmarks specifically designed for evaluating active perception.
To bridge this gap, we propose the first simulation benchmark specifically designed for evaluating active manipulation.
%


\section{Method}

We first formulate the active manipulation task (\cref{subsec: problem formulation}).
Then, we elaborate on {\ourmodel},  including its architecture (\cref{subsec: framework}) and training strategies (\cref{subsec: training pipeline}).
Finally, we describe the construction of {\ourdataset} dataset and details about {\oursimbenchmark} (\cref{subsec: dataset and benchmark}).

\begin{figure*}[t]
\centering
\includegraphics[width=\linewidth]{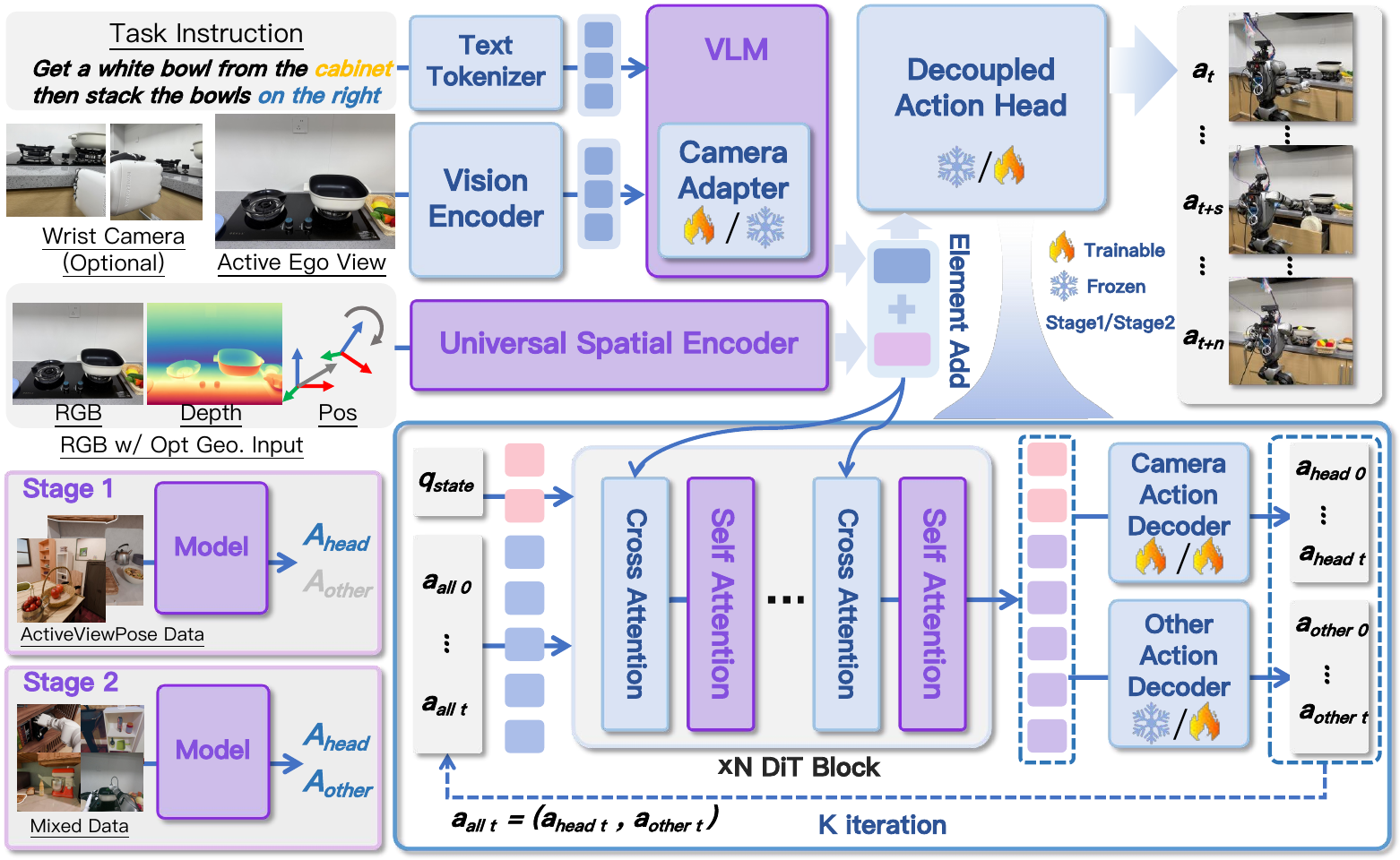}
\vspace{-6mm}
   \caption{\textbf{Overview of {\ourmodel}}. {\ourmodel} can process RGB images and task instructions and output camera movement and manipulation actions in a decoupled action space.
   This decoupled design enables the model to achieve active manipulation via a bottom-up, two-stage training strategy:
   First, large-scale embodiment-agnostic camera control data fosters \textit{semantic active perception}, which is encoded as prior knowledge in a camera adapter.
   Second, mixed data together with Universal Spatial Knowledge Injection flexibly incorporate various geometric configurations (\eg, absolute depth, camera intrinsics), thereby enhancing spatial precision for \textit{active-view execution}.}
\label{fig: pipline}
\vspace{-5mm}
\end{figure*}

\subsection{Problem Formulation}
\label{subsec: problem formulation}

We define the active manipulation task as learning a policy \( \pi_\theta: \mathcal{O} \times \mathcal{L} \rightarrow \mathcal{A} \). Given an observation \( O_t \in \mathcal{O} \) and a language instruction \( L \in \mathcal{L} \), the policy predicts a joint action trajectory \( A_t = \{A_{\text{head}, t}, A_{\text{other}, t}\} \in \mathcal{A} \). 
At each timestep \( t \), the observation \( O_t \) comprises the current RGB image \( I_t \in \mathbb{R}^{H \times W \times 3} \) and optional 3D geometric information \( G_t \) (\eg, depth maps and camera intrinsics/extrinsics). 
To ensure temporal consistency and smooth execution, we adopt an action chunking strategy where the policy predicts an action sequence over a horizon \( k \). The head camera action sequence is defined as \( A_{\text{head}, t} = \{a_{\text{head}}^\tau\}_{\tau=t}^{t+k-1} \), where each \( a_{\text{head}}^\tau \in \mathbb{R}^2 \) represents the relative pitch and yaw adjustments of the robot's head. 
Similarly, the manipulation action sequence is denoted as \( A_{\text{other}, t} = \{a_{\text{other}}^\tau\}_{\tau=t}^{t+k-1} \). 
Specifically, for our platform—a Unitree G1 humanoid equipped with dual 7-DoF arms and dual 6-DoF Inspire hands—each \( a_{\text{other}}^\tau \in \mathbb{R}^{26} \) represents the 26-DoF joint position deltas at timestep \( \tau \), specifying the relative angular change for each controllable joint.

Unlike prior works that unify camera motion and manipulation into a single action space, we decouple them and propose a two-stage learning strategy for active manipulation.
This approach first enrich the model with semantic active perception priors in a data-efficient way, and further optimize both camera control and manipulation actions for effective active-view execution in bottom-up manner.
%
%
Moreover, our observation space includes diverse 3D geometric inputs, enabling the model to acquire strong spatial understanding and more resilient to viewpoint changes.
Therefore, our model excels in complex active manipulation tasks with strong semantic-aware perception.


\begin{figure*}[t]
\centering
\includegraphics[width=\linewidth]{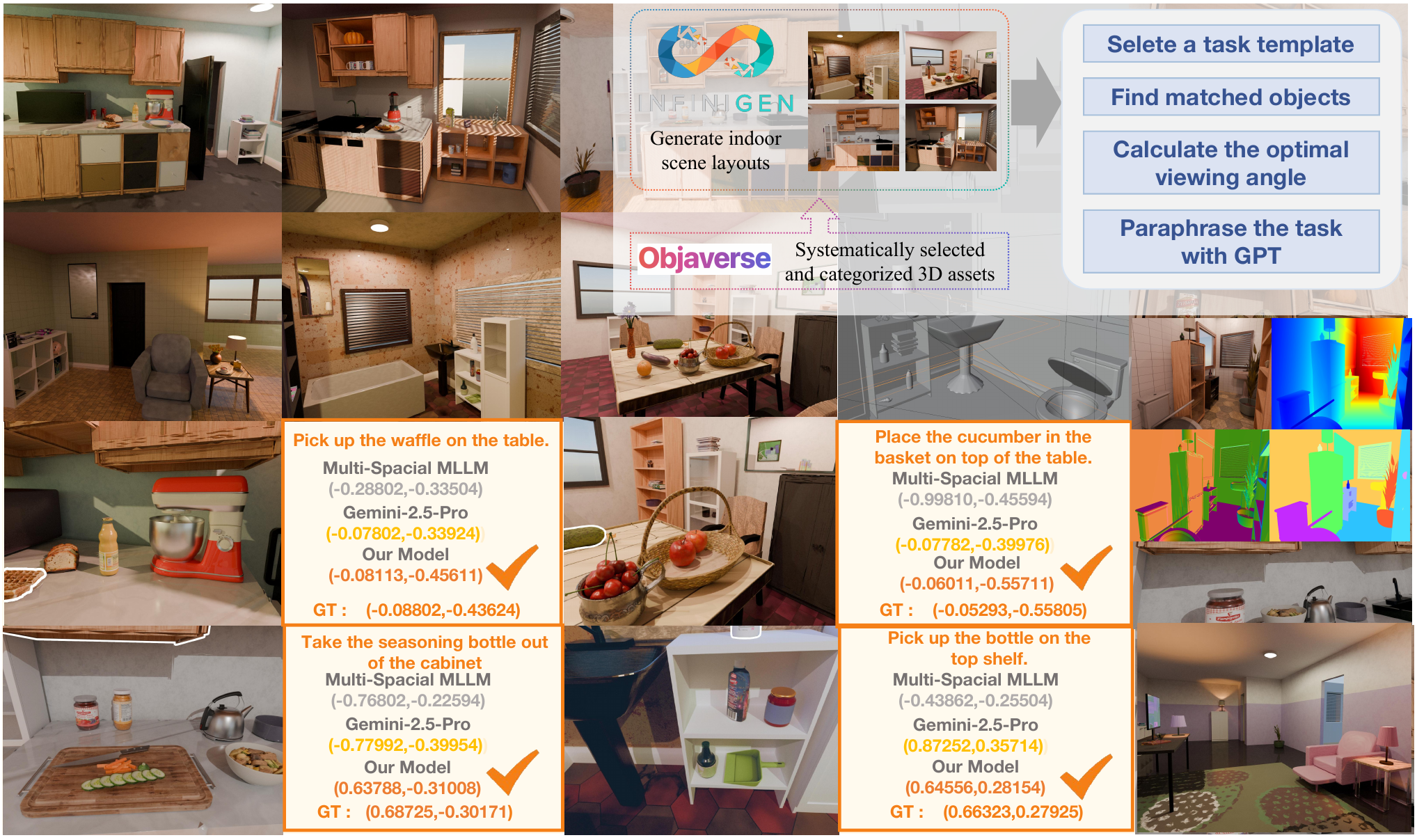}
\vspace{-6mm}
   \caption{\textbf{Overview of ActiveViewPose-200K}. It is a high-quality dataset comprising 200k image-language and camera movement pairs, enriched with highly detailed semantic annotations to enable semantic camera movement learning.
   }
\label{fig: dataset}
\vspace{-5mm}
\end{figure*}

\subsection{Architecture}
\label{subsec: framework}

Effective active manipulation requires both semantic active perception and robust active-view execution.
However, we find that existing active manipulation approaches do not support semantic input~\cite{ViA,breyer2022closed,zhang2023affordance}. 
Moreover, a straightforward fine-tuning of existing VLA models~\cite{pi0,gr00t} is insufficient. 
First, directly adding camera movement into the existing VLA action space would break the large-scale fixed-view manipulation priors learned from previous training.
Second, this disruption to such priors increases data demands, making it difficult to scale up due to the high cost and time-consuming nature of real-world data collection~\cite{open-television,activeumi,yu2025egomi}.
Therefore, we propose \textit{Decoupled Action Heads and Camera Adapter} to enable our model to acquire rich semantic active perception priors and retain general manipulation knowledge in a data-efficient manner.
Meanwhile, existing VLA methods often exhibit insufficient view-invariant 3D geometric consistency, hindering their ability to generate temporally stable action chunks under an actively changing egocentric perspective.
To bridge this gap, we propose \textit{Universal Spatial Knowledge Injection}, which efficiently leverages as much 3D information as possible to directly optimize the action output.
%

\vspace{+1mm}
\noindent \textbf{Decoupled Action Heads and Camera Adapter.}
We design two parts.
\textbf{(1)} Camera Adapter: We add a Camera Adapter using LoRA~\cite{hu2022lora} on the VLM to learn semantic active perception priors without altering the original VLM’s weights.
It is trained on a large-scale semantic camera movement dataset (see \cref{subsec: dataset and benchmark}) to enhance the model’s semantic active perception capabilities.
\textbf{(2)} Decoupled Action Head:
In \cref{fig: pipline}, we design two decoders to decouple camera movement from other actions. This simple, lightweight decoupling approach enables the model to quickly learn accurate camera movements and other actions.

\vspace{+1mm}
\noindent\textbf{Universal Spatial Knowledge Injection.}
A most straightforward approach to enhance VLA’s 3D spatial perception and understanding is to introduce as much accurate 3D information as possible.
Therefore, we propose a Universal Spatial Knowledge Injection approach that supports the optional inclusion of arbitrary varieties of 3D geometry information and injects them into the action head to guide the action-decoding process.
As shown in \cref{fig: pipline}, we adopt a Universal Spatial Encoder inherited from a powerful feed-forward 3D geometry model~\cite{keetha2025mapanything}. This encoder can support a wide range of 3D geometry configurations as inputs without requiring retraining or architectural modifications.
The encoded spatial tokens are element-wise added to the VLM output tokens, and then the mixed tokens are injected into the Decoupled Action Head during the action denoising process.
In this way, it not only strengthens semantic camera movement but also enhances active-view execution.

\subsection{Two-Stage Training Strategy}
\label{subsec: training pipeline}
We design a two-stage training strategy to equip our model with semantic active perception and active-view execution to achieve active manipulation in a bottom-up manner.

\vspace{+1mm}
\noindent\textbf{Stage 1: Semantic Active Perception Alignment.}
Our goal is to first equip the model with sufficient semantic active perception priors, which requires a large amount of high-quality data. 
However, existing datasets are all collected from a fixed near-optimal view. 
To fill this gap, we propose a large-scale, high-quality dataset, {\ourdataset}, comprising 200k image-language and camera movement pairs (see \cref{subsec: dataset and benchmark}).
We thus use this dataset to train Camera Adapter and Camera Action Decoder by supervising camera movement (see \cref{fig: pipline}).
The objective is to minimize the Mean Squared Error between the predicted ego camera movement $A_{\text{head}}$ and the ground-truth $A_{\text{head,t}}^*$, defined as 
$
\mathcal{L}_{\text{stage1}} = \mathcal{L}_{\text{MSE}}(A_{\text{head,t}},\, A_{\text{head,t}}^*).
$
Upon completion, the model develops a robust prior for semantic active perception.

\vspace{+1mm}
\noindent\textbf{Stage 2: Active Manipulation Fine-tuning.}
Building upon the semantic active perception prior, this stage trains the model to perform effective active manipulation. 
The data is a mixture of {\ourdataset} and active manipulation robot data (See \cref{subsec: dataset and benchmark}). 
We freeze the Camera Adapter and train the Decoupled Action Head using MSE loss, such that 
$
\mathcal{L}_{\text{stage2}} = \lambda_{\text{head}}\,\mathcal{L}_{\text{head}} 
+ \lambda_{\text{other}}\,\mathcal{L}_{\text{other}},
$
Therefore, the model acquires comprehensive active manipulation skills.
More details can be found in the Appx.~\ref{suppsec: DaPaVe_architecture}.

\subsection{ActiveViewPose-200K and ActiveManip-Bench}
\label{subsec: dataset and benchmark}

\textbf{ActiveViewPose-200K Dataset.}
Existing large-scale robotics datasets~\cite{open-x,wu2024robomind} have been instrumental in advancing policy learning. However, they are predominantly composed of data collected from nearly optimal fixed camera viewpoints.
To fill this gap, we introduce {\ourdataset}, the first large-scale, high-quality semantic active perception dataset, comprising 200k image-language and optimal camera movement pairs.
We construct this dataset through an efficient semi-automatic process. 
First, we meticulously collected 4k high-quality, semantically annotated assets from Objaverse~\cite{deitke2023objaverse} and collected 500 diverse scenes.
Next, we use a heuristic algorithm to produce a large number of image-to-camera movement pairs. 
Meanwhile, we semi-automatically built 3,000 detailed task templates. These templates, along with the images, are sent to GPT-4o to generate relevant instructions, which are then manually refined.
This data collection process is highly efficient and cost-effective, and can be used to train and evaluate the model's semantic active perception capability.

\begin{figure*}[t]
\centering
\includegraphics[width=\linewidth]{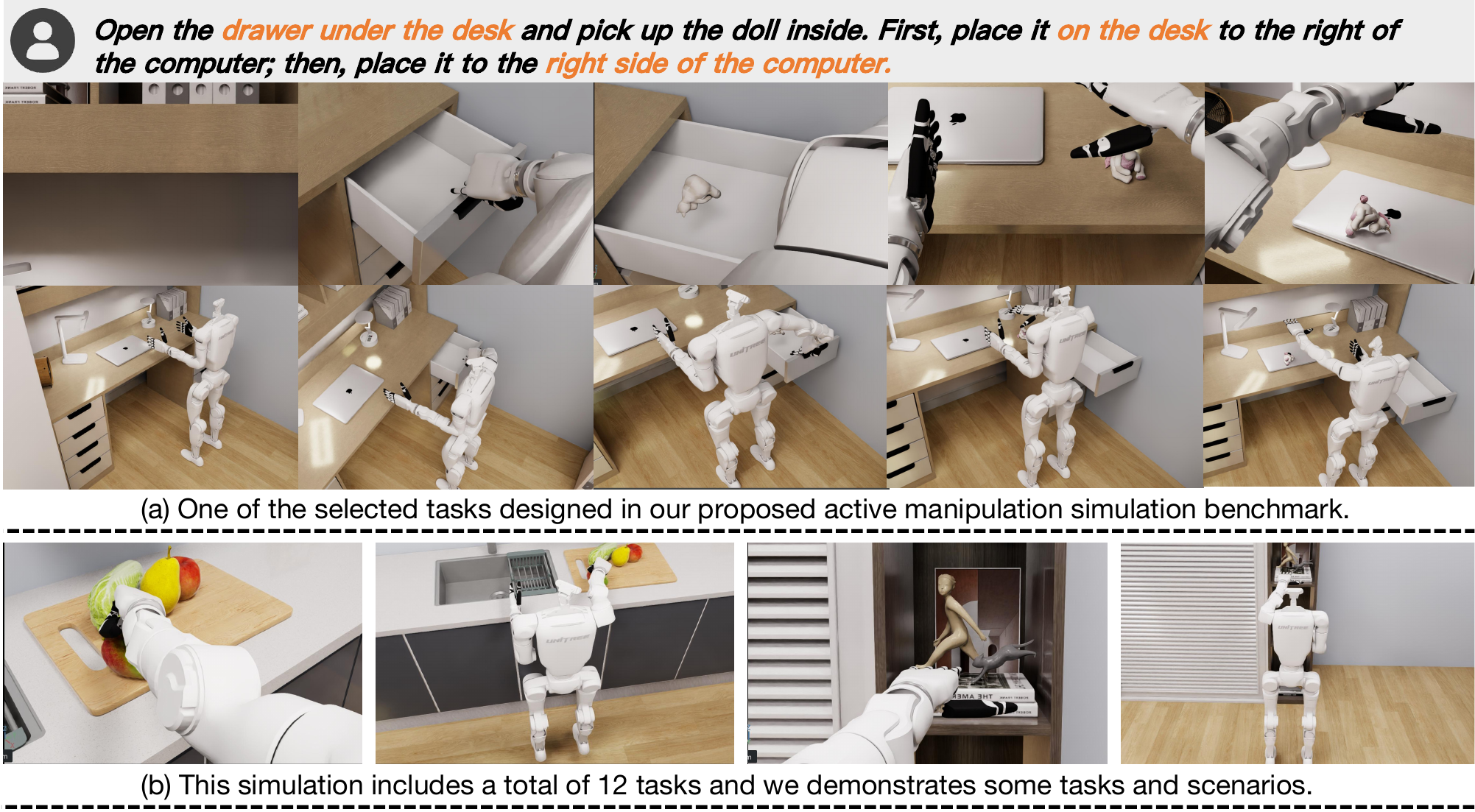}
\vspace{-6mm}
\caption{\textbf{Overview of {\oursimbenchmark}}: It is the first simulation benchmark to evaluate active manipulation beyond traditional fixed-view settings. {\oursimbenchmark} features 12 richly annotated tasks across 100 objects and 20 diverse scenes.}
\label{fig: sim_bench}
\vspace{-5mm}
\end{figure*}

\vspace{+1mm}
\noindent\textbf{\oursimbenchmark.}
Following the bottom-up strategy, after we equip our model with sufficient semantic active perception priors, we need a benchmark for both training and evaluating our model's active manipulation capability.
However, existing simulation benchmarks~\cite{xiang2020sapien,liu2023libero,mees2022calvin} are all confined to a fixed view and do not evaluate this capability.
To meet our requirements, we are the first to propose an Active Manipulation benchmark named {\oursimbenchmark}.
{\oursimbenchmark} is built on NVIDIA Isaac Sim and features a G1 humanoid equipped with a pair of Inspire Hands and an active, head-mounted perception camera. 
{\oursimbenchmark} now covers 100 different objects distributed across 20 distinct scenes, featuring 12 semantic active manipulation tasks with rich and detailed annotations.
It is also designed to be easily extensible, allowing for the simple addition of new scenes and objects.
%
More details can be found in the Appx.~\ref{suppsec: simbenchmark}.

\section{Experiments}
\label{sec:experiments}

\begin{table}[t]
\centering
\small
\caption{The performance of semantic active perception evaluation. We report the success rate (\%) compared to current general VLMs and specialized spatial VLMs.}
\label{tab:perception_eval}
\begin{tabular}{lcccc}
\toprule
\textbf{Method}    & \textbf{Val}  & \textbf{Test1} & \textbf{Test2} & \textbf{Avg.} \\
\midrule
Qwen2.5-VL-72B~\cite{bai2025qwen2}     & 63.9          & 65.1           & 58.0           & 62.3          \\
Multi-SpatialMLLM~\cite{xu2025multi}       & 72.8          & 74.3           & 63.6           & 70.2          \\
Gemini-2.5-Pro~\cite{comanici2025gemini}     & 73.3          & 76.5           & 68.2           & 72.7          \\
\midrule
\textbf{Ours (Stage 1)} & \textbf{85.5} & \textbf{89.1}  & \textbf{78.3}  & \textbf{84.3} \\
\bottomrule
\end{tabular}
\vspace{-3mm}
\end{table}

In this section, we conduct a series of experiments to rigorously evaluate our proposed framework. We want to answer the following key questions:
\textbf{(1)} How much semantic active manipulation priors can our model actually acquire (\cref{ssec:eval_perception}) ?
\textbf{(2)} Where and to what extent does our active camera design improve upon previous fixed-camera or wrist-camera setups (\cref{ssec:eval_sim})?  
\textbf{(3)} Compared to existing VLA models, how does our model architecture enhance active manipulation capabilities (\cref{ssec:eval_real})?
\textbf{(4)} How well does our model generalize to out-of-distribution (OOD) scenarios  (\cref{ssec:genernalize})?
\textbf{(5)} What role do each of the system components play in enhancing its overall performance (\cref{ssec:ablation})?

\begin{table*}[t]
\caption{Evaluation results for fixed and dynamic cameras in simulation of {\oursimbenchmark}. We report the success rate (\%) compare to different camera configurations with the same architecture. P.a.P and A.M denote Pick-and-Place and articulated Manipulation. }
\vspace{-2mm}
\centering
\scriptsize
\begin{tabular}{l|ccccccc}
\bottomrule[1pt]
Method & Unoccluded P.a.P & Occluded P.a.P & Out-of-View P.a.P & Unoccluded A.M & Occluded A.M & Out-of-View A.M & Avg. \\
\midrule
Fixed Camera                   & 74                         & 46                   & 11                         & 52                     & 27       & 7       &36.17   \\
Fixed Camera + Wrist Camera    & 83                         & 62                    & 28                          & 66                     & 51           &24     &52.33     \\
Active Camera + Wrist Camera   & \textbf{86}                & 75                    & 70                          & 74                    & 68         & 66      &73.16      \\ 
\midrule
\textbf{Ours (Active Camera)}  & 85                        & \textbf{78}           & \textbf{72}                 & \textbf{76}            & \textbf{70}  & \textbf{68} & \textbf{74.83}\\
\hline 
\end{tabular}
\label{tab: simulation_exp}
\vspace{-4mm}
\end{table*}

\subsection{Experimental Setup}
\label{ssec:exp_setup}

\noindent \textbf{Datasets and Benchmarks.}
We design a series of experiments leveraging different components of our dataset and benchmarks:
\textbf{(1)} For the first experiment to evaluate semantic active perception (\cref{ssec:eval_perception}), we test models on our {\ourdataset}, which is split into Train/Val/Test1/Test2. 
In Test1, the language instructions contain explicit positional indicators (\eg, ``on the left''), whereas in Test2, the instructions omit detailed camera movements, requiring the model to infer them from both the text and the image.
\textbf{(2)} For the second experiment to evaluate the effect of fixed or dynamic cameras across different types (\cref{ssec:eval_sim}), we assess performance on {\oursimbenchmark} (see \cref{subsec: dataset and benchmark}), which comprises 6 task types: Unoccluded/Occluded/Out-of-View Pick-and-Place, as well as Unoccluded/Occluded/Out-of-View Articulated Manipulation.
\textbf{(3)} For All subsequent experiments (\cref{ssec:eval_sim}, \cref{ssec:eval_real}, \cref{ssec:ablation}), we employ our real-robot teleoperated dataset, including 4 task categories: Occluded/Out-of-View Pick-and-Place, and Occluded/Out-of-View Articulated Manipulation.
%
See Appx.~\ref{suppsec: real_world_exp} for details.


\vspace{+1mm}
\noindent \textbf{Baselines.}
We compare our method against several strong baselines:
\textbf{(1)} In the first experiment, we compare our model with current powerful VLM, including the general models (\eg, Qwen2.5-VL-72B~\cite{bai2025qwen2}, Gemini-2.5-Pro~\cite{comanici2025gemini}), as well as the specialized spatial VLM (\eg, Multi-SpacialMLLM~\cite{xu2025multi}). 
\textbf{(2)} In the second experiment, we adopt the same underlying architecture but explore different camera configurations:
(a) Fixed Camera: A standard VLA model that operates from a static camera—this is the most common setup in prior work. We use our model but keep the head camera fixed.  
(b) Fixed Camera + Wrist Camera: We again use our model with a fixed head camera and introduce two wrist cameras.  
(c) Active Camera + Wrist Camera: We use our model, combining the active head camera with two wrist cameras.
\textbf{(3)} Finally, in the third experiment, we compare our model with existing VLA baselines, $\pi_0$~\cite{pi0} and GR00T-N1~\cite{gr00t}, both of which are recognized for their strong general-purpose capabilities.

\vspace{+1mm}
\noindent \textbf{Evaluation Metrics.} 
 For all experiments, we report the success rate. For the first experiment, a prediction is considered successful if it falls within a tolerance of the ground-truth pitch and yaw changes (See Appx.~\ref{suppsec: discussion}).

\begin{table}[t]
\caption{Performance on active manipulation in real-world settings. We report the success rate (\%) compared to the existing VLA models. Our approach achieves the best performance.}

\vspace{-1mm}
\centering
\scriptsize
\setlength{\tabcolsep}{2pt}
\begin{tabular}{l|ccccc}
\toprule
\multirow{2}{*}{Method} & Occluded & Out-of-View & Occluded & Out-of-View & \multirow{2}{*}{Avg.}\\
                        & Pick-and-Place & Pick-and-Place & Arti-Manip  & Arti-Manip\\
\midrule
$\pi_0$~\cite{pi0}   & 55 & 45 & 45 & 35 & 45.00\\
GR00T-N1~\cite{gr00t} & 60 & 55 & 50 & 50 & 53.75\\
\midrule 
\textbf{Ours}  & \textbf{90} & \textbf{85} & \textbf{85} & \textbf{80} & \textbf{85.00}\\
\bottomrule
\end{tabular}
\label{tab:real-world}
\vspace{-4mm}
\end{table}

\subsection{Semantic Active Perception Evaluation}
\label{ssec:eval_perception}
We evaluate our model on the {\ourdataset} test sets and present the following findings.
%
As shown in Table~\ref{tab:perception_eval}, our model significantly outperforms powerful VLMs across all test splits, especially on test2, where semantic understanding is paramount. Notably, our model achieves an average score 11.6\% higher than Gemini-2.5-Pro~\cite{comanici2025gemini} with only 2B parameters.
This is because the generalist VLMs struggle to translate abstract instructions into precise camera poses, whereas our specialized training yields a much more accurate and robust perception policy.
It demonstrates that acquiring an embodied skill like semantic active perception is not an emergent property of generalist VLMs. Instead, direct training on specialized data, as enabled by our ActiveViewPose dataset, is crucial for learning a functional and precise policy.

\subsection{Fixed and Dynamic Cameras Evaluation.}
\label{ssec:eval_sim}

We conduct this evaluation using {\oursimbenchmark} in simulation. 
The following paragraphs present our analyses.

\vspace{+1mm}
\noindent \textbf{Dynamic viewpoints are crucial in active manipulation.} In \cref{tab: simulation_exp}, under a fixed viewpoint, the success rates of unoccluded, occluded, and out-of-view tasks all decrease substantially—especially for the Out-of-View task, which drops by more than 60\%. 
This result indicates that a fixed camera greatly limits the model’s ability to explore the accessible space, leading to failures for active manipulation.

\vspace{+1mm}
\noindent \textbf{The combination of Fixed camera and Wrist camera remains insufficient for active manipulation.}
We observe that, although this setup outperforms using a single fixed camera setting, it still falls short of active camera setting in every task—especially out-of-view tasks, which heavily depend on the model’s ability to rotate the main viewpoint for active perception of the environment. Since the fixed main viewpoint can never observe all objects, this combination fails to address active manipulation tasks.

\vspace{+1mm}
\noindent \textbf{The gain from combining an active camera with a wrist camera is minimal and can even degrade certain performance metrics.} This outcome may seem counterintuitive, as one would expect more viewpoints to yield better results. However, we find that if the model already possesses robust dynamic perception (\ie, the ability to avoid occlusions and observe effectively) and strong 3D spatial understanding, adding additional dynamic viewpoints does not necessarily improve performance in active manipulation tasks.

\subsection{Comparison with existing VLA models}
\label{ssec:eval_real}
We compare our model with existing VLAs whose action spaces are extended to incorporate camera movement. Both are fine-tuned for active manipulation tasks.

\vspace{+1mm}
\noindent \textbf{Directly fine-tuning existing VLA models is insufficient to fully address active manipulation tasks}
In \cref{tab:real-world}, both GR00T-N1 and $\pi_0$ underperform our approach in nearly all tasks. Two main factors account for this shortfall:  
\textbf{(1)} Direct VLA fine-tuning does not provide sufficient active perception priors.  
\textbf{(2)} Current VLA models lack dedicated modules for enhancing active-view manipulation.  
These observations validate the necessity of our work.

\vspace{+1mm}
\noindent \textbf{Our bottom-up strategy proves to be highly effective.} In \cref{tab:real-world}, our bottom-up approach significantly outperforms direct VLA fine-tuning, demonstrating that first establishing robust semantic active perception priors and then learning active manipulation tasks is extremely effective. Meanwhile, the decoupled action head benefits more fully from this two-stage training strategy, thereby enhancing the model’s active manipulation capability.



\subsection{Generalization Ability Evaluation.}
\label{ssec:genernalize}
We evaluate our model’s generalization ability in two real-world tasks, examining variations in lighting, objects, and operational backgrounds.
As shown in \cref{tab:generalization}, our model demonstrates strong generalization to previously unseen objects, indicating robust high-level semantic understanding that enables it to interpret out-of-distribution objects and correctly follow instructions. Meanwhile, across different lighting conditions and scenarios, the model continues to exhibit strong generalization, suggesting that incorporating 3D information bolsters its 3D spatial perception and ensures reliably robust operations in diverse environments.

\begin{table}[t]
\caption{ Performance on generalization ability evaluation. We report the success rate (\%). Our model demonstrates robust generalization when performing active manipulation across unseen objects, varying lighting conditions, and diverse scenes.}
\label{tab:generalization}
\vspace{-1mm}
\centering
\scriptsize
\setlength{\tabcolsep}{1.5pt} 
\begin{tabular}{l *{7}{c}}
\toprule
\textbf{Task Name} & \textbf{Object 1} & \textbf{Object 2} & \textbf{Light 1} & \textbf{Light 2} & \textbf{Scene 1} & \textbf{Scene 2} & \textbf{Original} \\
\midrule
Occluded        & \multirow{2}{*}{85} & \multirow{2}{*}{90}& \multirow{2}{*}{90}& \multirow{2}{*}{95}& \multirow{2}{*}{90}& \multirow{2}{*}{85}& \multirow{2}{*}{\textbf{90}} \\
Pick-and-Place  &&&&&&\\
\midrule
Out-of-View        & \multirow{2}{*}{85} & \multirow{2}{*}{90}& \multirow{2}{*}{80}& \multirow{2}{*}{90}& \multirow{2}{*}{85}& \multirow{2}{*}{85}& \multirow{2}{*}{\textbf{85}} \\
Pick-and-Place  &&&&&&\\
\midrule
Occluded       & \multirow{2}{*}{80} & \multirow{2}{*}{85}& \multirow{2}{*}{80}& \multirow{2}{*}{85}& \multirow{2}{*}{85}& \multirow{2}{*}{80}& \multirow{2}{*}{\textbf{85}} \\
Arti-Manip  &&&&&&\\
\midrule
Out-of-View        & \multirow{2}{*}{75} & \multirow{2}{*}{80}& \multirow{2}{*}{80}& \multirow{2}{*}{85}& \multirow{2}{*}{80}& \multirow{2}{*}{75}& \multirow{2}{*}{\textbf{80}} \\
Arti-Manip  &&&&&&\\
\bottomrule
\end{tabular}
\vspace{-4mm}
\end{table}

\subsection{Ablation Studies}
\label{ssec:ablation}
We conduct a series of ablation experiments on 4 real-world tasks to evaluate the effectiveness of different components in our method. We present our analyses below.

\vspace{+1mm}
\noindent\textbf{The bottom-up training strategy is crucial for active manipulation tasks.} \cref{tab: ablation} indicates that both training stages substantially improve the model’s performance. In particular, for out-of-view tasks, omitting Stage 1 drastically reduces the success rate—by half in artifact manipulation—showing that first equipping the model with semantic active perception priors is essential for completing active manipulation tasks. Meanwhile, omitting Stage 2 reduces the model’s overall success rate, underscoring the necessity of active manipulation finetuning.

\vspace{+1mm}
\noindent\textbf{The decoupled formulation is correct.} We find that using a unified single action decoder for simultaneously learning camera movement and other actions, compared to using a decoupled approach, weakens the model’s active manipulation ability. This is mainly because camera movement is inherently embodiment-agnostic. Forcing the use of a unified action decoder couples the two training stages in the action space, not only disrupting the semantic active perception priors but also impairing the active manipulation capability.

\vspace{+1mm}
\noindent\textbf{Using a lightweight camera adapter to learn camera movement proves superior to fully finetuning.} As shown in \cref{tab: ablation}, comparing the Camera Adapter approach with fully finetuning reveals that fully finetuning leads to performance drops across all four experiments. This suggests that the camera adapter can learn camera rotation while preserving the VLM’s general high-level semantic information, a crucial factor for maintaining the model’s fundamental understanding of active manipulation tasks.

\vspace{+1mm}
\noindent\textbf{Universal Spatial Knowledge Injection greatly enhances the model’s robustness for basic operations under active views}. In \cref{tab: ablation}, we observe that omitting Universal Spatial Knowledge Injection leads to a 15\% performance drop even in relatively simple Occluded Pick-and-Place tasks. This indicates that maintaining a consistent 3D spatial understanding is critical for active manipulation. It further underscores how Universal Spatial Knowledge Injection effectively guides the model’s action output, since prior 3D knowledge can facilitate implicit 3D learning.

\begin{table}[t]
\caption{Ablation Study on the effect about training strategy of Stage 1 and Stage2, decoupled action head (D.A.H.), camera adapter (C.A.), and universal spatial knowledge injection (U.S.K.I). We report the avarage success rate (\%).}
\vspace{-2mm}
\centering
\scriptsize
\setlength{\tabcolsep}{2pt}
\begin{tabular}{l|ccccc}
\bottomrule[1pt]
\multirow{2}{*}{Ablation}                  & Occluded & Out-of-View & Occluded & Out-of-View & \multirow{2}{*}{Avg.}\\
                                           & Pick-and-Place & Pick-and-Place & Arti-Manip  & Arti-Manip\\
\midrule
w/o. Stage 1 & 65 & 55 & 50 & 45 & 53.75\\
w/o. Stage 2 & 75 & 60 & 70 & 60 & 66.25\\
w/o. D.A.H. & 80 & 70 & 70 & 65 & 71.25\\
w/o. C.A. & 80 & 75 & 70 & 70 & 73.75\\
w/o. U.S.K.I. & 75 & 75 & 65 & 60 & 68.75\\
\hline 
\end{tabular}
\label{tab: ablation}
\vspace{-4mm}
\end{table}

\section{Conclusion}
\label{sec: conclusion}

In this paper, we propose {\ourmodel}, an end-to-end framework that unifies semantic active perception with active-view execution for active manipulation.
By first leveraging large-scale semantic camera control data, we use a decoupled action space to learn semantic active perception, acquired as prior knowledge in a camera adapter. 
We then integrate Universal Spatial Knowledge Injection to handle arbitrary geometric inputs, improving active-view execution. 
Moreover, we present {\ourdataset}, a large-scale dataset for semantic camera learning, and {\oursimbenchmark}, a tailored benchmark for active manipulation. 
Extensive experiments show the effectiveness of {\ourmodel} and highlight its potential for human-like robots.

\section*{Acknowledgement}
This work was supported by the National Natural Science Foundation of China (62476011), and by the Beijing Natural Science Foundation (L252060).
We sincerely thank Yuran Wang and Xinqiang Yu for their insightful discussions and valuable feedback regarding the development of our simulation environment. We are also grateful to Yibo Li for refining the teaser figure, and to Zhengliang Cai, Ning Chen, Yangyang Wei and Yankai Fu for their indispensable support with the real-world experiments, particularly in troubleshooting and repairing the robot hardware.

{
    \small
    \bibliographystyle{ieeenat_fullname}
    \bibliography{main}
}


\clearpage
\appendix
\setcounter{page}{1}
\maketitlesupplementary

\noindent In the supplementary document, we will explain in detail the parts that are not fully elaborated upon in the main text due to page limitations. These explanations are organized as follows:
\newline
\begin{itemize}

    \item Sec.~\ref{suppsec: discussion}: Discussions on certain details within the paper.
    \newline
    \item Sec.~\ref{suppsec: implementation_of_dataset}: Details of {\ourdataset}, including data filtering, collection, and QA generation. 
    \newline
    \item Sec.~\ref{suppsec: simbenchmark}: Implementation details of {\oursimbenchmark}, including environment setup, task design and data collection methods.
    \newline
    \item Sec.~\ref{suppsec: real_world_exp}: More Details on real world experimental settings, including hardware setup, task protocols and generalization testing.
    \newline
    \item Sec.~\ref{suppsec: DaPaVe_architecture}: Implementation details of {\ourmodel}, including architecture and training details of each stage.
    \newline
    \item Sec.~\ref{suppsec: more demonstrations}: More Demonstrations of {\ourmodel}.
    \newline
    \item Sec.~\ref{suppsec: limitation}: Discussion on Limitations and Future Work.
\end{itemize}

\section{Discussion}
\label{suppsec: discussion}
%
%

\begin{figure*}[]
\centering
\includegraphics[width=\linewidth]{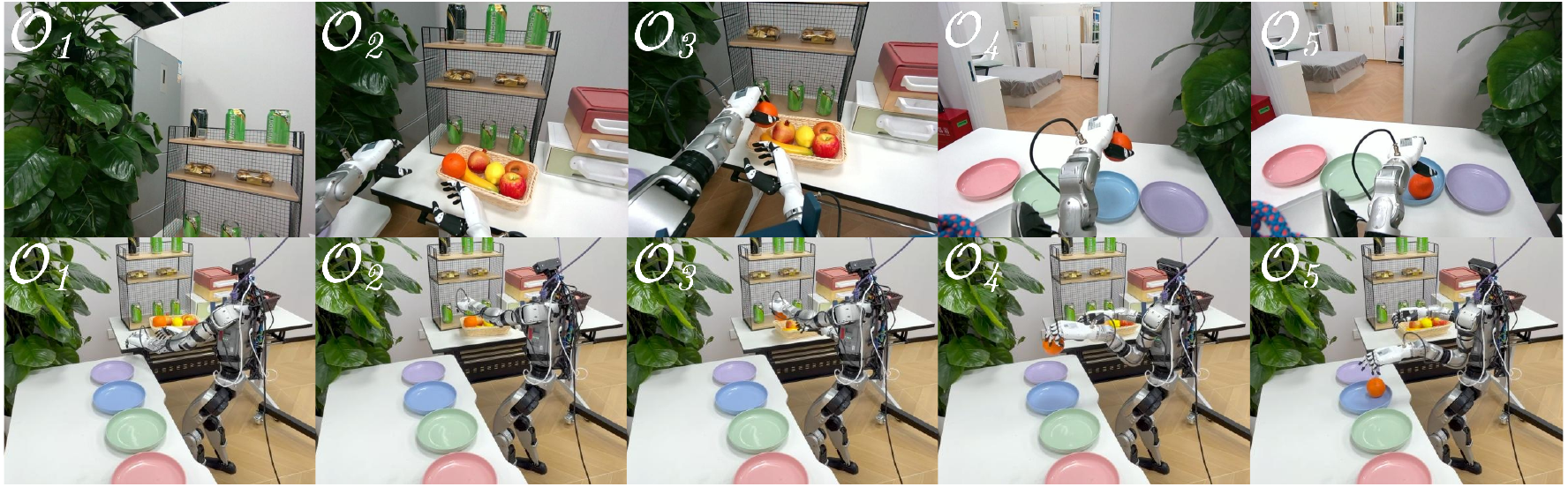}
\vspace{-5mm}
   \caption{Real-world Execution roll-outs (ego \& third view).}
\label{fig: rebuttal}
\vspace{-5mm}
\end{figure*}

\noindent \textbf{View Variation.}
We add tasks with larger camera movements and conduct real-world experiments following previous evaluation protocol.
\cref{fig: rebuttal} shows that execution roll-outs involve large viewpoint shifts ($>90^\circ$).
Only our method succeeds with 45\% success rate, validating the effectiveness of our active-view strategy and model design.

\noindent \textbf{Semantic Active Perception Evaluation.}
The values in \cref{fig: dataset} represent the final pitch and yaw changes of the camera movement.
We detail the inference of different models and the evaluation process as follows:
\textbf{(1)} Multi-SpatialMLLM~\cite{xu2025multi} directly predicts the total pitch and yaw changes, leveraging its strong camera understanding.
\textbf{(2)} Gemini-2.5-Pro~\cite{comanici2025gemini} predicts the center point of the target view after camera movement, and then computes the corresponding pitch and yaw changes based on camera geometry, as it has strong pointing capability.  
\textbf{(3)} Our model directly outputs a camera movement chunk ($A_{head}$), and we obtain the final pitch and yaw changes by accumulating the deltas across chunks.
For evaluation, a prediction is considered successful if it falls within a predefined tolerance of the ground-truth pitch and yaw changes; otherwise, it fails.

\vspace{+1mm}
\noindent \textbf{Additional Wrist Cameras.}
We attribute the performance drop in active perception settings with wrist cameras mainly to \textbf{\textit{Noise Information}} and \textbf{\textit{Data Mismatch}}.
\textbf{(1) \textit{Wrist cameras are not inherently detrimental}}:
They are useful when the ego (head) view is fixed.
In Tab.~2, Fixed + Wrist outperforms Fixed alone, especially for out-of-view settings, showing that wrist views offer valuable complementary information when head movement is restricted.
\textbf{(2) \textit{Noise Information}}: The head camera alone usually provides sufficient visual input for task completion, and in practice, teleoperators mainly rely on it to collect data.
While wrist cameras provide finer localization, especially for distant or small objects, they also introduce noise (\eg, occlusions) in active manipulation, which negatively impacts performance.
\textbf{(3) \textit{Data Mismatch}}:
The noise information actually further stems from data mismatch, as {\ourdataset} contains only head camera movements for Stages 1 and 2, while the paired head–wrist data is limited (~20k samples) and used only in Stage 2.
Although we try our best to decouple the head camera action space to specifically enhance active view capability, the scarcity of paired data limits effective learning of how to integrate the information from both head and wrist cameras.
We will scale up paired head–wrist data to improve active manipulation.

\section{\ourdataset}
\label{suppsec: implementation_of_dataset}
To train the model’s semantic active perception capability, we constructed a large-scale synthetic dataset---\ourdataset. This section details the four-stage pipeline: 3D asset curation, procedural scene synthesis, task-driven view annotation, and instruction augmentation.

\subsection{3D Asset Curation and Filtering}
We source 3D objects from the Objaverse dataset\cite{deitke2023objaverse}, utilizing LVIS annotations for semantic categorization. To ensure the objects are suitable for physical interaction simulation and geometric estimation, we applied a rigorous filtering protocol. The selection criteria are as follows:

\vspace{+1mm}
\noindent \textbf{Semantic Relevance}: 
Objects are restricted to common indoor categories (\eg, household items, tools, decorations).

\vspace{+1mm}
\noindent \textbf{Object Integrity}: 
We strictly select single objects rather than object assemblies or pre-arranged scenes to ensure flexible placement.

\vspace{+1mm}
\noindent \textbf{Geometric Stability}: 
Objects must possess a flat base to ensure stable placement on planar surfaces (\eg, tables, counters).

\vspace{+1mm}
\noindent \textbf{Visual Consistency}: 
Assets must exhibit realistic textures and color styles compatible with synthetic indoor environments.

\vspace{+1mm}
\noindent \textbf{Topology Cleaning}: 
We filter out objects with extraneous ``loose'' parts that affect bounding box calculation and collision size, such as wires attached to mice or cables connected to webcams.

\begin{figure*}[t]
\centering
\includegraphics[width=\linewidth]{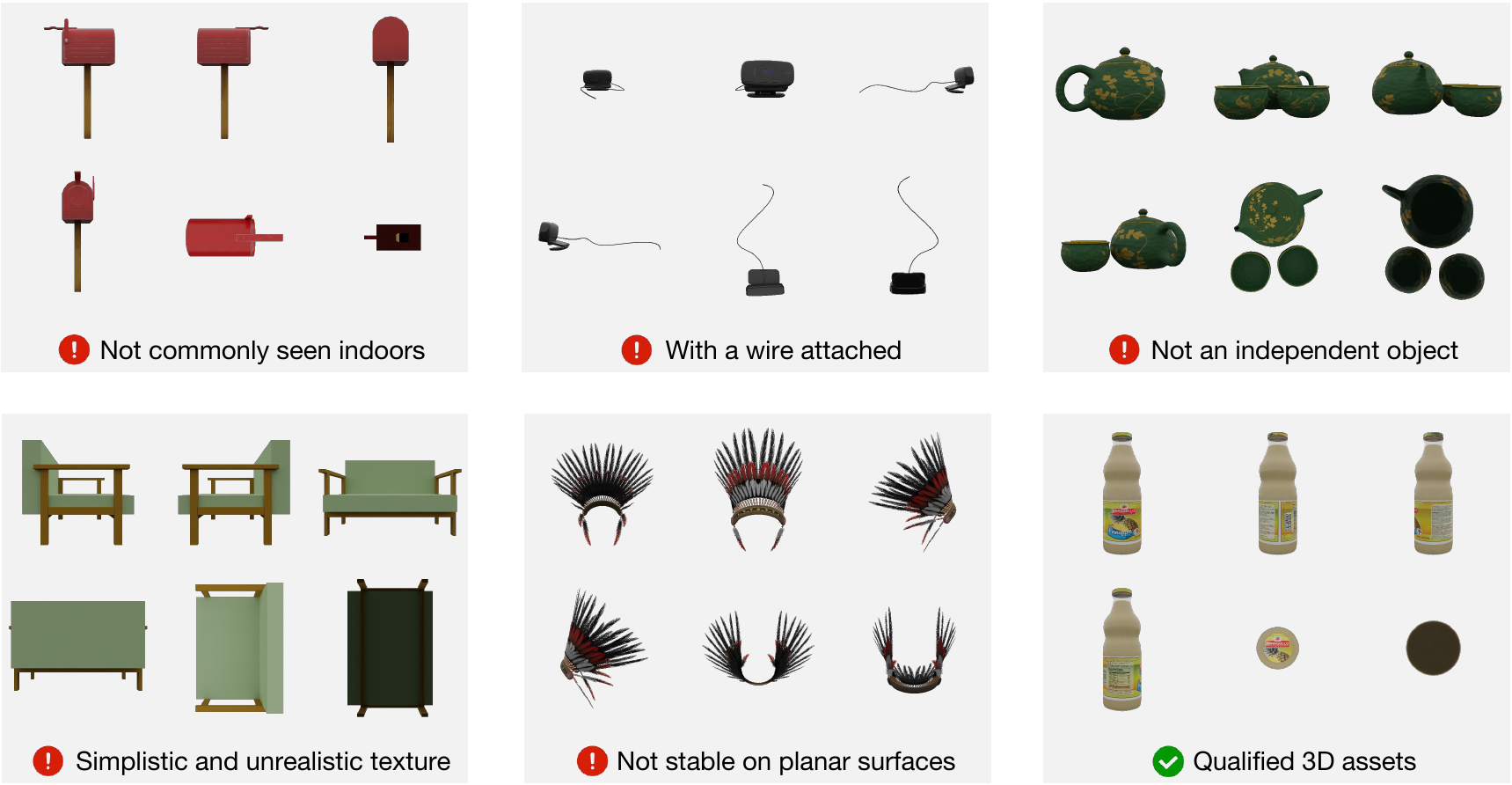}
   \caption{Examples of the asset curation process.
   }
\label{fig: asset}
\end{figure*}

\subsection{Procedural Scene Generation}
\begin{figure*}[t]
\centering
\includegraphics[width=\linewidth]{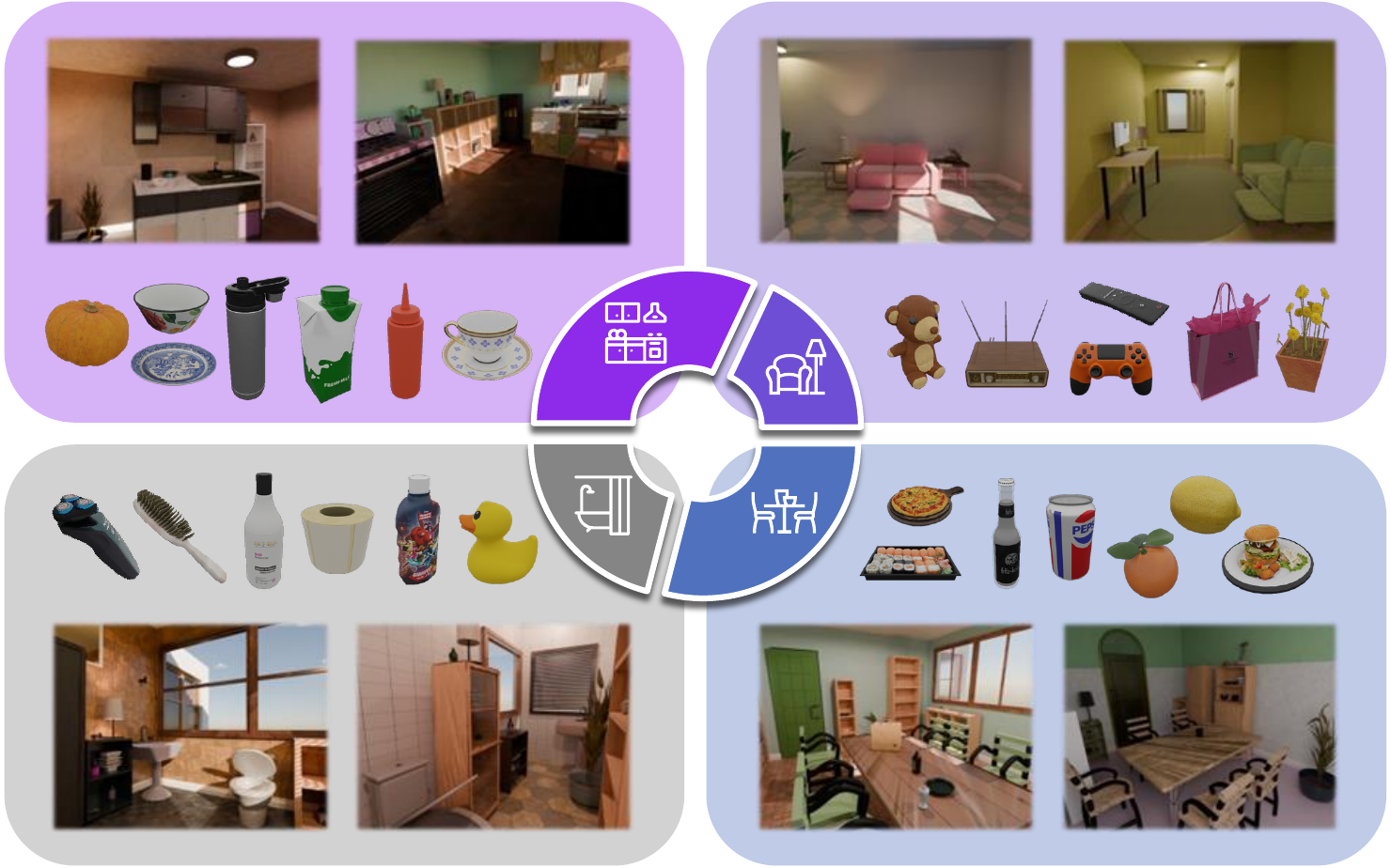}
   \caption{Generated synthetic environments using Infinigen integrated with our curated asset library. The scenes feature varying lighting conditions, furniture layouts, and clutter distributions.
   }
\label{fig: scene_examples}
\end{figure*}

We utilize \textit{Infinigen}\cite{raistrick2024infinigen}, a procedural scene generation framework, to construct diverse indoor environments. Our curated assets are integrated into the Infinigen asset library, replacing or augmenting default placeholders. 
To simulate real-world distribution, we generate scenes across four functional room types with specific probability distributions based on common household activity frequency: Kitchen ($32\%$), Living Room ($18\%$), Dining Room ($29\%$), and Bathroom ($21\%$). Figure \ref{fig: scene_examples} demonstrates the diversity of the generated environments.

\subsection{Task Formulation}
The robotic tasks are defined by a combination of atomic actions and specific prompt modalities that guide the camera movement strategies.

\vspace{+1mm}
\noindent \textbf{Atomic Actions.}
Our dataset covers a range of manipulation-oriented atomic tasks: \texttt{pick}, \texttt{reorient}, \texttt{open}, \texttt{close}, \texttt{pour}, and \texttt{place}.

\vspace{+1mm}
\noindent \textbf{Prompt Modalities.}
We design three distinct types of textual prompts to train the model's capability to interpret different forms of spatial instructions:

\begin{itemize}
    \item \textbf{Visual Centering:} The target is partially visible or off-center. The goal is to center the view for operation. \\
    \textit{Example:} ``Pick up/Reorient the [Object]. (the object is in the image)''
    \item \textbf{Spatial Directive:} The target is not currently visible. The instruction provides explicit relative direction. \\
    \textit{Example:} ``Turn left and pick up the [Object] on the table'' or ``Look up and retrieve the [Object] on the shelf.''
    \item \textbf{Common Sense:} The target is not visible, and no explicit direction is given. The agent must infer the location based on semantic context. \\
    \textit{Example:} ``Pick up the [Object] in the base cabinet'' (implying a downward look).
\end{itemize}

\vspace{+1mm}
\noindent \textbf{Data Augmentation.}
To enhance the model's capabilities, we introduce complex scenarios:
\begin{itemize}
    \item \textbf{Conditional Reasoning:} Referring to objects by attributes such as position sequences, size, or color. \\
    \textit{Example:} ``Pick up the second plate from left to right'' or ``Place the biggest apple inside the right basket.''
    \item \textbf{Container Interaction:} Tasks that require opening a container before grasping an item. \\
    \textit{Example:} ``Grasp the jar in the left drawer.''
\end{itemize}

\begin{figure*}[t]
\centering
\includegraphics[width=\linewidth]{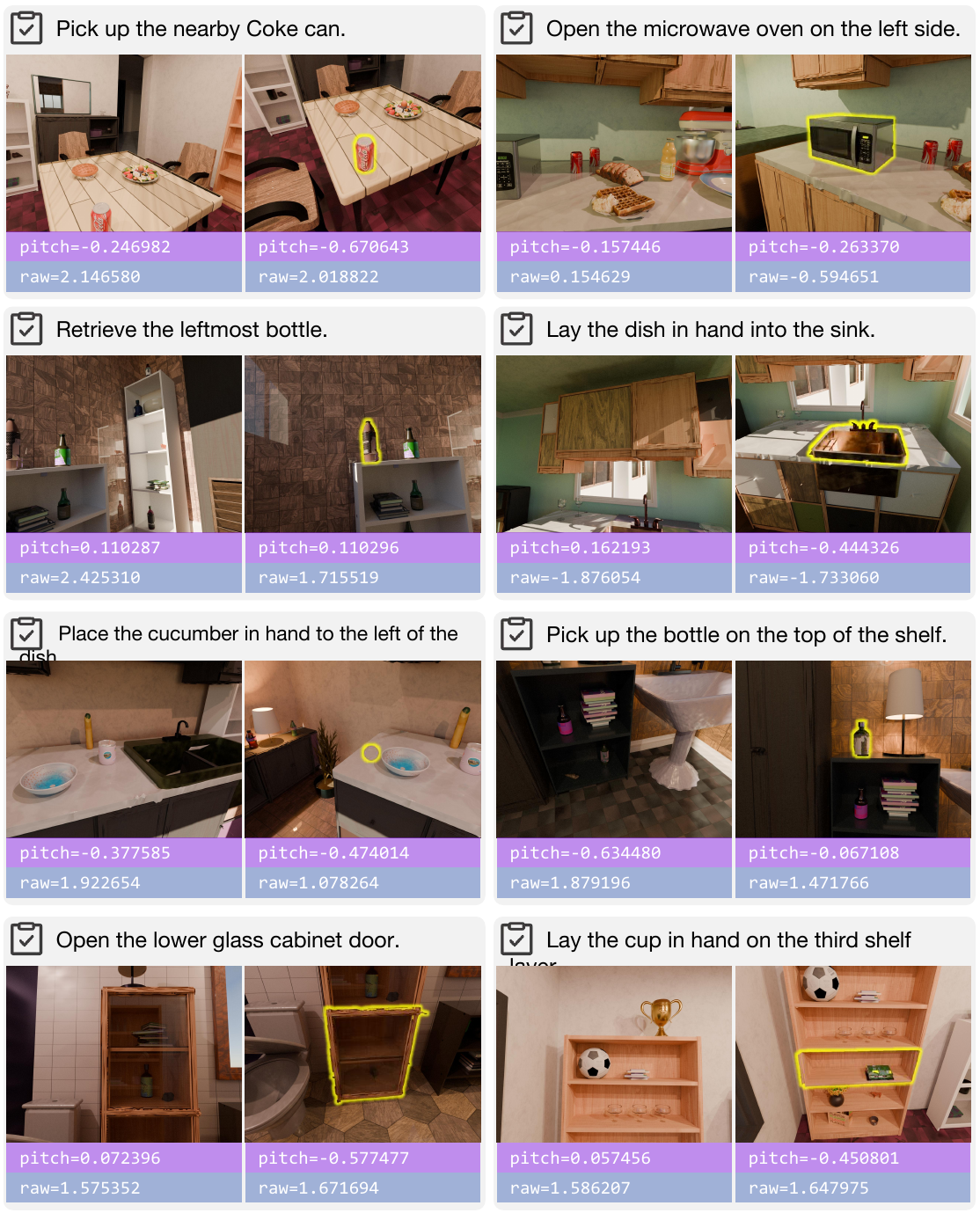}
\vspace{-6mm}
   \caption{Atomic Scene Tasks and Viewpoint Movement Examples.
   }
\label{fig: data_sing}
\vspace{-5mm}
\end{figure*}

\begin{figure*}[t]
\centering
\includegraphics[width=\linewidth]{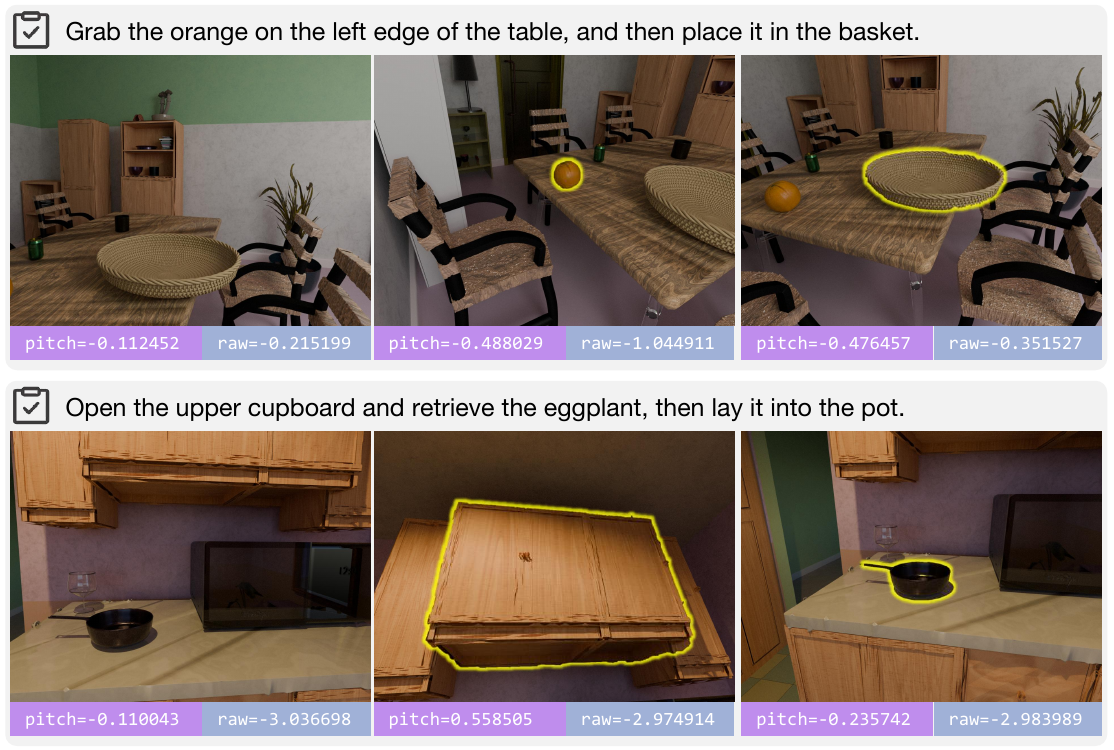}
\vspace{-6mm}
   \caption{Two-stage scene task and viewpoint movement example.
   }
\label{fig: data_mult}
\vspace{-5mm}
\end{figure*}

\noindent For the generated scenes, we randomly place objects within them and, according to different task templates, search for objects or object groups that satisfy the required positional relationships. We then sample a camera pose within a reasonable range and compute the optimal viewpoint for completing the task, which serves as the target view. This target view is subsequently perturbed within a prescribed range to produce the initial view. Representative samples are shown in Figure \ref{fig: data_sing} and Figure \ref{fig: data_mult}.

\subsection{Instruction Augmentation via LLM}
To prevent the model from overfitting to rigid template patterns, we employ GPT-4 to rewrite the template-generated instructions. We prompt the LLM to introduce linguistic diversity while preserving the core semantic meaning and spatial relationships. The prompt design encourages the inclusion of diverse cues, such as explicit directional commands or implicit commonsense references (e.g., implying "cupboard" is "above").
The prompt structure used for instruction rewriting is shown in Listing \ref{lst:gpt_prompt}.

\begin{lstlisting}[basicstyle=\ttfamily\footnotesize, backgroundcolor=\color{myblue!50}, caption={Prompt used for diverse instruction generation.}, captionpos=t, breaklines=true, label={lst:gpt_prompt}]
You are a task-description rewriting expert for robot learning, whose goal is to enrich data diversity by converting structured template instructions into more natural and varied expressions.

Rewriting Rules:

Mandatory Principles:
1. Maintain imperative form: every rewritten instruction must be a direct imperative that states the action to be performed.
2. Preserve object names exactly: names must be kept without synonym substitution or generalization.
3. Keep the action equivalent: although wording may vary, the intended action must remain functionally the same.
4. Do not alter task difficulty: rewriting must not introduce new objects or change the actual complexity.

Acceptable Variations:
1. Verb diversity: choose more specific or natural verbs suited to the object and target location.
   - verbs should align with functional use of the object and destination.

2. Sentence-structure changes: reorganize the sentence.
   - Convert simple to complex by adding purpose or manner adverbials.

3. Contextual rationale: add reasons or purposes for the action.
   - Leverage common uses of the objects to supply functional descriptions.

4. Manner modifiers: include relevant adverbs.

5. Preparatory phrases: occasionally insert expressions implying prior steps.

Prohibited Changes:
1. Altering object names.
2. Changing target locations or target objects.
3. Replacing core referents with synonyms.
4. Introducing new objects or locations.
5. Switching to non-imperative forms (e.g., "You should...", "It is necessary to...").
6. Adding unrelated task steps.

Task:
Original instruction: {INSTRUCTION}

Provide only the rewritten instruction.
\end{lstlisting}

\section{\oursimbenchmark}
\label{suppsec: simbenchmark}
In this section, we provide a detailed description of how {\oursimbenchmark}  
is constructed, including the simulation environment details, simulation task design and efficient data collection methods. 

\subsection{Simulation Environment Details}
\vspace{+1mm}
\noindent \textbf{Simulation Platform.} 
We build {\oursimbenchmark} environment on the NVIDIA’s Isaac Sim platform\cite{isaac}, a dedicated robotics simulation application that delivers photorealistic rendering and physically accurate simulations for robotics research and development.As illustrated in Figure \ref{fig: supp_simbenchmark_1}, {\oursimbenchmark} provides photorealistic rendering and physically accurate simulations.

\vspace{+1mm}
\noindent \textbf{Scene Configuration.} 
In {\oursimbenchmark}, we include 100 different objects and 20 distinct scenes. These simulation assets are high-quality digital-twin indoor-scene articulated objects with precise collision meshes and dynamic joint parameters, ensuring that simulated motion faithfully replicates real-world physics and kinematics. We collect our simulation assets from \cite{artvip}. Similarly, we carefully curated a collection of high-quality, large-scale indoor environments from \cite{3d-front,artvip}. Additionally, the asset format supports quick scene and object substitution through simple operations, allowing users to tailor environments to their individual requirements.

\vspace{+1mm}
\noindent \textbf{Detailed Robot Specifications} 
In our simulation, we employ a G1 robot equipped with two Inspire third-generation dexterous hands, each featuring 6 degrees of freedom, while each arm provides 7 degrees of freedom. We have replaced the robot’s original fixed head with a 2-DoF active head, and we use this newly designed system to execute tasks. Through our teleoperation algorithm or a deployed model, we can directly control the 7 joints of both arms, the 6 joints of each dexterous hand, and the 2 joints of the head.

\vspace{+1mm}
\noindent \textbf{Camera Viewpoints.}
Our simulator supports three types of camera viewpoints:  
(1) An active ego-camera viewpoint.  
(2) Wrist-camera viewpoints for the robot’s left and right arms.  
(3) A fixed third-person camera viewpoint.Each viewpoint has a default RGB input resolution of 224×224, although users can flexibly adjust this resolution as needed.

\subsection{Active Manipulation Task Suite.}

We have designed 12 Active Manipulation Tasks with comprehensive semantic annotations. To rigorously evaluate robotic capabilities, we structure our task suite along two orthogonal dimensions: \textbf{(1) Task Objectives}, ranging from atomic skills to long-horizon reasoning, and \textbf{(2) Visual Complexity}, ranging from fully visible scenarios to those requiring active semantic exploration.

\subsubsection{Hierarchical Task Objectives}

We classify the 12 tasks into three categories based on their horizon length and semantic dependencies. Table \ref{tab:spp_simbench_task_overview} presents the success criteria, while Fig. \ref{fig: supp_simbenchmark_2}, \ref{fig: supp_simbenchmark_1} provides detailed visualizations.

\begin{itemize}
    \item \textbf{Atomic Manipulation Tasks.} 
    We include six fundamental tasks focusing on rigid body manipulation and articulated object interaction. 
    In {Pick} and {Reorient}, the robot must grasp an object to achieve a target 6-DoF pose. Specifically, {Pick} focuses on lifting the object to a specified height, while {Reorient} requires rotating the object to a target orientation from an arbitrary initial state.
    For the articulated four tasks \{{Open}, {Close}\}\{{Drawer}, {Cabinet}\}, the goal value specifies the geometric state of the articulated joint. Specifically, success for {Drawer} tasks depends on the linear distance of prismatic joints, whereas {Cabinet} tasks involve the angular state of revolute joints. The initial state is randomized to ensure the agent learns active adaptation rather than memorizing trajectories.

    \item \textbf{Short-Horizon Composite Tasks.}
    Building upon atomic skills, we introduce three tasks that require sequential execution: {PickAndPlace}, {OpenCloseDrawer}, and {OpenCloseCabinet}.
    In {PickAndPlace}, the agent must establish a stable grasp, transport the object avoiding collisions, and release it at a precise target location. 
    The tasks $\{{OpenClose}\}\{{Drawer}, {Cabinet}\}$ serve as a testbed for reversible manipulation, requiring the robot to drastically alter the environment state (opening) and then restore it (closing), necessitating robust motion planning and continuous state tracking.

    \item \textbf{Long-Horizon Sequential Tasks.}
    We introduce three high-complexity tasks that embody rich semantic content: {FetchFromDrawer}, {FetchFromCabinet}, and {PourLiquid}.
    
    The {FetchFrom} tasks mimic realistic household scenarios comprising a four-stage chain: \textit{Open-Pick-Place-Close}. Unlike simple picking, the robot must first manipulate the container (drawer/cabinet) to reveal the workspace, then perform confined-space manipulation to grasp the target object inside the occluded area, place it at a designated goal, and finally close the container to complete the task cycle. This imposes strict preconditions where each action's success depends on the geometric state resulting from the previous step.
    
    Finally, distinct from rigid-body tasks, {PourLiquid} challenges the robot's ability to manipulate dynamic fluids. The robot must pick up a bottle containing liquid and tilt it at a precise angle to transfer the content into a target container, requiring reasoning about liquid dynamics and volume retention.
\end{itemize}

\begin{table*}[htbp]

    \centering
    \caption{Overview of the 12 Active Manipulation Tasks. The tasks are categorized by horizon length and complexity. Success is rigorously defined by geometric thresholds (position, rotation, joint state) or physical quantities (liquid volume), requiring the robot to maintain the goal state for a stabilization period (e.g., 2 seconds).}
    \label{tab:spp_simbench_task_overview}
    \resizebox{\textwidth}{!}{
    \begin{tabular}{l l l}
        \toprule
        \textbf{Category} & \textbf{Task Name} & \textbf{Success Criteria (Goal State)} \\
        \midrule
        
        \multirow{6}{*}{\textbf{Atomic}} 
        & \textsc{Pick} & Object height form table $> 5$cm; velocity $\approx 0$. \\
        & \textsc{Reorient} & Orientation error relative to target $< 15^\circ$. \\
        & \textsc{OpenDrawer} & Prismatic joint distance (opening) $> 90\%$ of analytical limit. \\
        & \textsc{CloseDrawer} & Prismatic joint distance (opening) $< 5\%$ (fully closed). \\
        & \textsc{OpenCabinet} & Revolute joint angle (opening) $> 80^\circ$. \\
        & \textsc{CloseCabinet} & Revolute joint angle (opening) $< 5^\circ$. \\
        \midrule
        
        \multirow{3}{*}{\textbf{Short-Horizon}} 
        & \textsc{PickAndPlace} & Object position error $< 5$cm at target; stable placement. \\
        & \textsc{OpenCloseDrawer} & \textit{Seq:} Fully open ($>90\%$) $\to$ Fully closed ($<5\%$). \\
        & \textsc{OpenCloseCabinet} & \textit{Seq:} Fully open ($>80^\circ$) $\to$ Fully closed ($<5^\circ$). \\
        \midrule
        
        \multirow{5}{*}{\textbf{Long-Horizon}} 
        & \multirow{2}{*}{\textsc{FetchFromDrawer}} & \textit{Seq:} Open Drawer $\to$ Pick Object $\to$ Place on Counter \\
        & & $\to$ Close Drawer. (All preconditions must be met). \\
        \cmidrule(l){2-3}
        & \multirow{2}{*}{\textsc{FetchFromCabinet}} & \textit{Seq:} Open Cabinet $\to$ Pick Object $\to$ Place on Counter \\
        & & $\to$ Close Cabinet. (All preconditions must be met). \\
        \cmidrule(l){2-3}
        & \textsc{PourLiquid} & Liquid in target container $> 80\%$ of source volume; Spilled $< 10\%$. \\
        
        \bottomrule
    \end{tabular}
    }
\end{table*}

\subsubsection{Visual Complexity and Initialization}

To evaluate the robustness of agents against visual uncertainties and their ability to utilize language for active perception, we introduce diversity in the initial object placements independent of the task type. For each task, the target object's initial state is sampled from three levels of visual complexity:

\begin{itemize}
    \item \textbf{Unoccluded (Fully Visible).} 
    The target object is initialized at a random position within the robot's reachable workspace and remains fully contained within the camera's field of view (FoV). Although the object is clearly visible, the agent must handle significant variations in the object's 6-DoF pose and background clutter, requiring robust pose estimation and grounding capabilities.

    \item \textbf{Occluded (Partially Observable).} 
    This category challenges the agent with incomplete visual information, presenting two distinct scenarios: 
    (1) \textit{Visual Truncation}, where the target is located at the periphery of the camera frame, leaving only a fraction of the object visible; and 
    (2) \textit{Physical Occlusion}, where the target is blocked by extraneous objects in the environment. 
    Success in this mode requires strategic active behaviors rather than passive perception. Specifically, for truncated views, the agent must actively adjust its viewpoint to bring the complete object into the field of view. Conversely, when facing physical occlusion, the agent must possess the capability to interact with the environment—identifying and moving away the occluding obstacles to reveal the target object.

    \item \textbf{Out-of-View (Semantically Grounded Search).} 
    This represents the highest difficulty level, where the target object is completely absent from the initial observation. In this setting, the agent must rely on the \textit{semantic cues} embedded in the task instruction to locate the object. 
    For instance, given the instruction \textit{"pick the bottle from the bottom cabinet"}, the agent must infer that the target is likely situated in a lower spatial region. Consequently, it must perform active perception—moving its camera or base to explore the environment guided by linguistic priors—before it can commence the manipulation phase.
\end{itemize}

\subsection{Data Collection and Generation Pipeline}

To construct a large-scale, high-quality dataset for active manipulation, we implemented a three-stage pipeline encompassing human demonstration collection, automated simulation-based augmentation, and semantic instruction enrichment.

\subsubsection{VR-based Human Teleoperation.}
We developed a high-fidelity teleoperation system inspired by \cite{open-television} to capture natural human behaviors. The system utilizes the \textit{Apple Vision Pro} as the interface hardware. 
This setup allows for the holistic capture of the operator's intent: the headset tracks the 6-DoF head pose to render an ego-centric active perception viewpoint, while the hand tracking capabilities capture precision dexterity. 
The detected human wrist poses and finger states are retargeted in real-time to the simulated robot's end-effector pose and gripper state via analytical Inverse Kinematics (IK), ensuring that the collected trajectories are kinematically feasible for the robot.

\subsubsection{Scalable Data Augmentation.}
Relying solely on human teleoperation is labor-intensive. To scale up our dataset diversity, we employ two strategies:
\begin{itemize}
    \item \textbf{Visual and Asset Randomization:} Our simulation framework supports hot-swapping of object assets and background environments. We systematically randomize textures, lighting conditions, and distractors to enhance the visual robustness of the trained policies.
    \item \textbf{Self-Improving Trajectory Generation:} To scale the quantity of trajectory data while maintaining human-like behavioral patterns, we adopt the methodology proposed in DexFlyWheel \cite{dexflywheel}. Instead of simple replay, we utilize a hybrid framework combining Imitation Learning (IL) with Residual Reinforcement Learning (RL).
    Specifically, we first train a base policy using human demos. We then employ a self-improving flywheel mechanism where Residual RL explores local adjustments to correct failures in the original demos. This process efficiently generates diverse, high-quality successful trajectories that preserve the nuanced intent of the original human demonstrations, significantly alleviating the scarcity of dexterous manipulation data.
\end{itemize}

\subsubsection{Semi-Automatic Semantic Enrichment.}
To bridge the gap between rigid template commands and natural language variation, we propose a semi-automatic semantic augmentation pipeline. 
We first define structured logical templates for each task, such as:
\begin{center}
\texttt{pick the [object] at [position] and place to [target\_position]}
\end{center}
These templates are then fed into a Large Language Model (LLM) to generate linguistically diverse and spatially specific instructions. For example, the LLM might expand the template into:
\begin{quote}
    \textit{"Pick the blue pen located to the left of the computer and place it onto the notebook."}
\end{quote}
Finally, to ensure grounding accuracy, all generated instructions undergo a round of human verification to correct for potential hallucinations or spatial mismatches.

\section{Real-World Experiments}
\label{suppsec: real_world_exp}
In order to better evaluate the differences between our model and the baseline, and to further assess the model’s generalization and sim-to-real transferability, we conducted extensive experiments on a physical robot platform.
The experimental design is structured into three parts: hardware setup, task protocols, and generalization testing.

\subsection{Hardware Setup}
\begin{figure}[t]
\centering
\includegraphics[width=\linewidth]{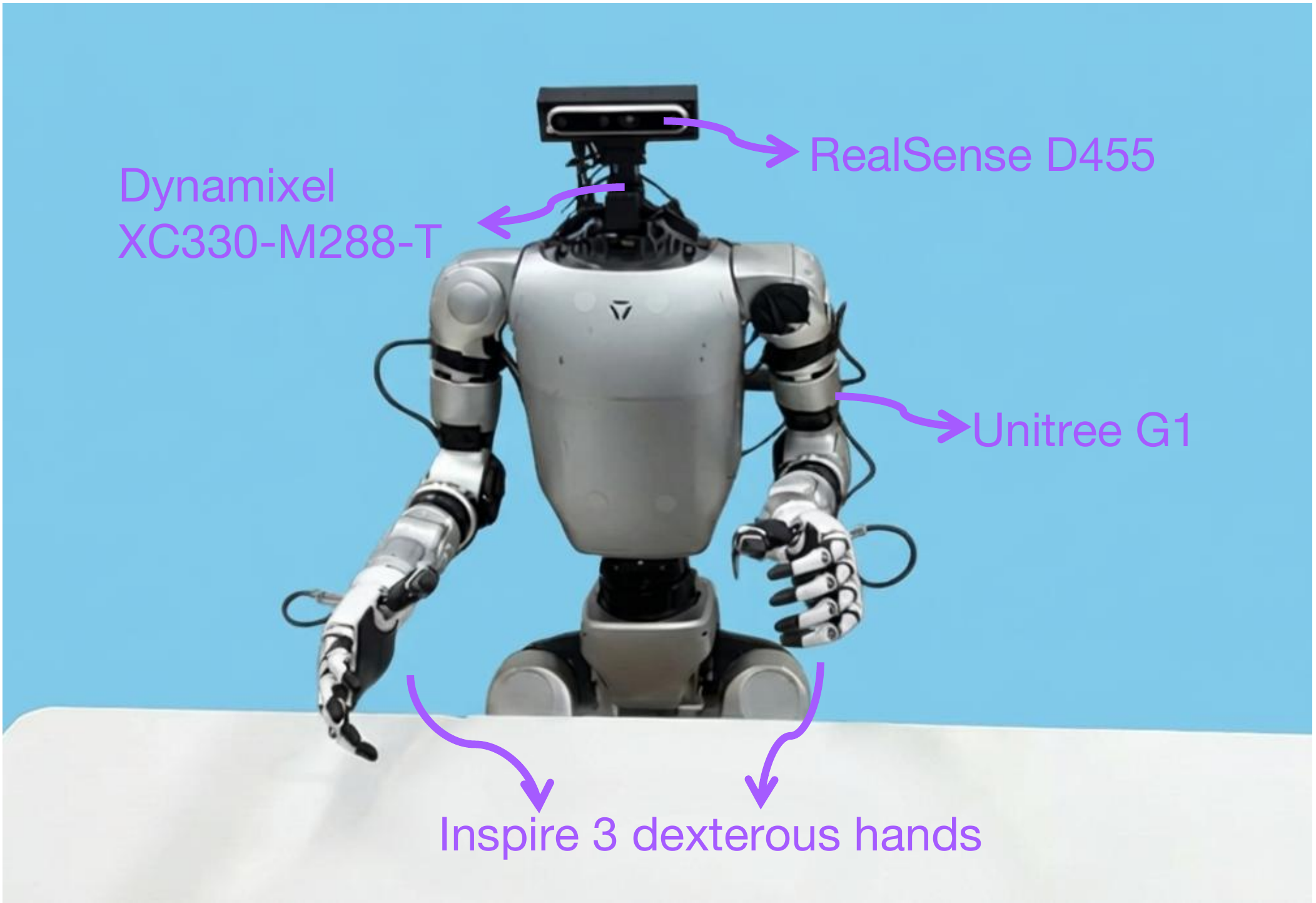}
\vspace{-6mm}
   \caption{We provide the type of the robot, dexterous hand, camera, and servos used in our real-robot experiments.
   }
\label{fig: real_hardware}
\vspace{-5mm}
\end{figure}

As illustrated in Figure \ref{fig: real_hardware}, our experimental platform is built upon the Unitree G1 humanoid robot. To enable high-fidelity dexterity, the robot is equipped with \textit{Inspire 3} dexterous hands. 
Crucially, to support active perception research, we developed a custom active head system. The head structure is fabricated using 3D-printed components and actuated by high-precision servos (Dynamixel XC330-M288-T), allowing for independent pitch and yaw movements.
A \textit{RealSense D455} RGB-D camera is mounted on this head unit, serving as the primary ego-centric sensor. This setup allows the agent to actively adjust its camera pose to explore the environment, mirroring the setup in our simulation.

\subsection{Real-World Task Protocols}

\begin{figure}[t]
\centering
\includegraphics[width=\linewidth]{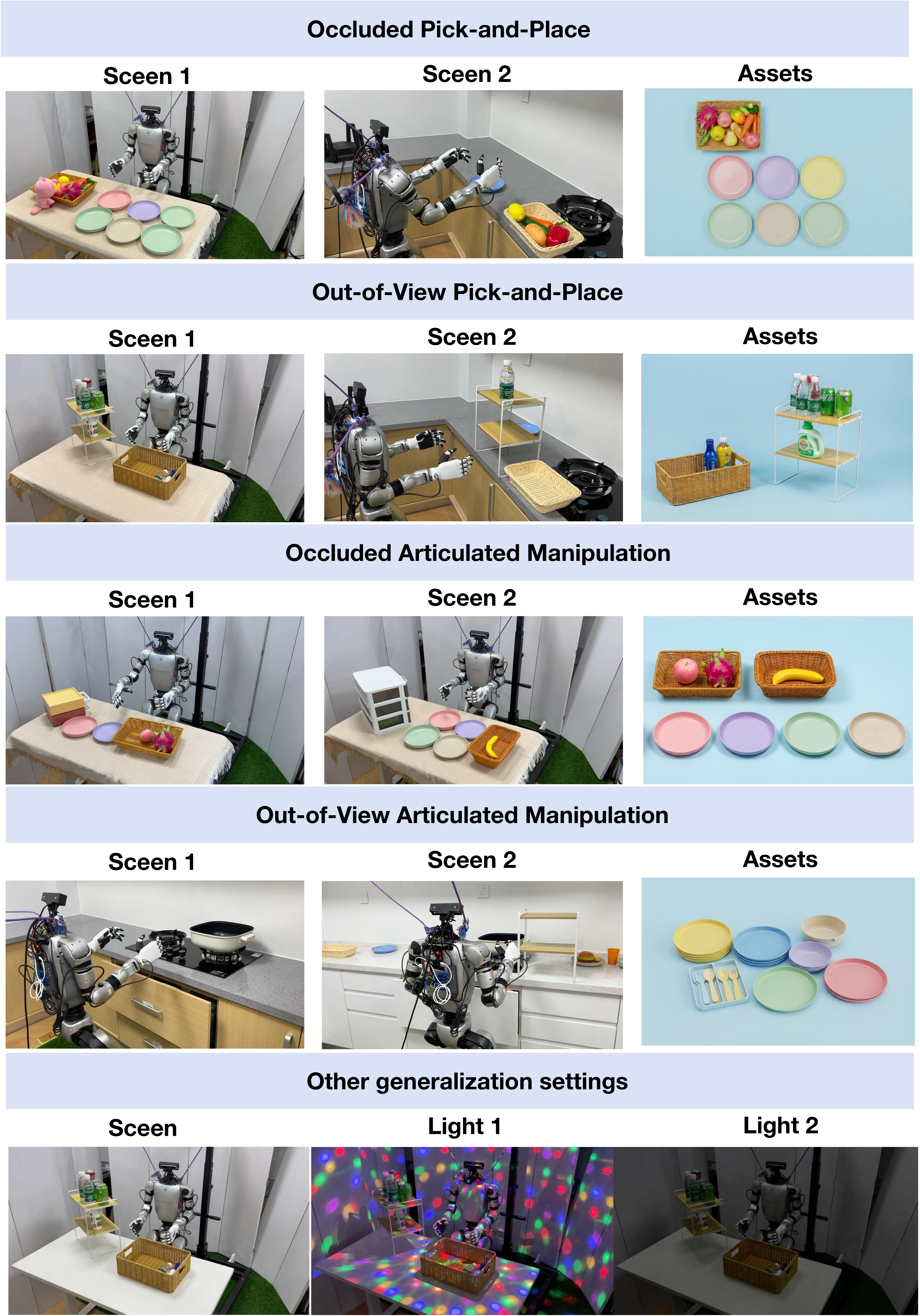}
\vspace{-6mm}
   \caption{This diagram illustrates the four tasks in our real-world experiment categories, along with the employed scenarios, assets, and varying lighting conditions.
   }
\label{fig: real_world_gen}
\end{figure}

We designed four distinct categories of tasks to evaluate the agent's manipulation and active perception capabilities under varying visibility constraints. The assets used in these experiments are shown in Fig.~\ref{fig: real_world_gen}.

\begin{enumerate}
    \item \textbf{Occluded Pick-and-Place.}
    \textit{Definition:} The target object or the placement goal is partially outside the initial camera field of view (FoV), or both are partially truncated. Success requires the agent to infer the full spatial information from partial observations and actively adjust its viewpoint to ensure precise manipulation.
    \textit{Instance:} The agent is tasked with transferring food items (e.g., fruits or bread) from a basket on the left side to a plate on the right side, where the containers are only partially visible initially.
    
    \item \textbf{Out-of-View Pick-and-Place.}
    \textit{Definition:} The target object or goal location is completely absent from the initial FoV. The agent must interpret semantic instructions (e.g., "on the left", "upper shelf") to actively rotate its head and locate the targets before manipulation.
    \textit{Instance:} The task involves retrieving a cup from a high shelf and placing it into a basket on the table's left side. The agent must utilize the semantic cue "upper shelf" to look up and locate the cup.

    \item \textbf{Occluded Articulated Manipulation.}
    \textit{Definition:} A long-horizon task where an articulated container (drawer) is partially visible. The agent must actively perceive to open the drawer, retrieve an object, close the drawer, and then actively search for the placement goal based on semantic cues.
    \textit{Instance:} The agent must identify a partially visible drawer on the left side of the desk, open it to retrieve an object, and subsequently place it into a basket on the right side.
    
    \item \textbf{Out-of-View Articulated Manipulation.}
    \textit{Definition:} This is the most challenging setting where the articulated container is completely out of sight initially. The agent relies entirely on contextual priors and language instruction (e.g., "cabinet under the counter") to guide its active exploration (e.g., looking down) to locate the handle, perform the manipulation chain (Open-Pick-Close), and locate the final placement goal.
    \textit{Instance:} We task the robot with retrieving a dish from a cabinet located underneath the countertop and placing it onto the tabletop. The agent must infer "under the counter" implies a downward gaze shift.
\end{enumerate}

\subsection{Generalization Evaluation}
To rigorously assess the robustness of our policy, we evaluate calibration in Out-of-Distribution (OOD) settings across three dimensions, as shown in Fig.~\ref{fig: real_world_gen}:
\begin{itemize}
    \item \textbf{Unseen Objects:} We select two novel objects and drawers that were never present during training to test geometric generalization.
    \item \textbf{Unseen Scenes:} The robot is deployed in a completely novel table setup and background arrangement distinct from the training environment.
    \item \textbf{Lighting Variation:} We introduce lighting conditions unseen in the dataset (e.g., dim evening light or strong directional sunlight) to evaluate visual robustness.
\end{itemize}

\section{SaPaVe Architecture and Implementation}
\label{suppsec: DaPaVe_architecture}
SaPaVe is constructed upon a modular foundation comprising a decoupled diffusion policy, a parameter-efficient vision-language adapter, and a universal spatial encoder. In this section, we detail the architectural innovations, training protocols, and evaluation metrics.

\subsubsection{Decoupled Action Heads via Diffusion Transformer}

For precise manipulation and active perception, we model the joint policy $\pi$ using a variant of the Diffusion Transformer (DiT) analogous to System 1 in GR00T N1. 
Given the current observation state $q_t$ and a target action chunk $A_t \in \mathbb{R}^{T \times D}$, the model learns to predict the noise $\epsilon$ added to the action latent.
However, unlike monolithic outputs, we propose a \textbf{Decoupled Action Head} architecture to handle the distinct dynamics of the active head and the robotic body.
The DiT backbone consists of alternating self-attention (handling action temporal dependencies) and cross-attention blocks (conditioning on multi-modal embeddings). At the final decoding stage, instead of a single projection, the latent features split into two specialized Multilayer Perceptron (MLP) decoders:
\begin{enumerate}
    \item \textbf{Camera Action Decoder:} Predicts the 2-DoF kinematics for the active head (pitch and yaw), denoted as $\hat{A}^{cam}_t \in \mathbb{R}^{T \times 2}$.
    \item \textbf{Manipulation Action Decoder:} Predicts the 26-DoF joint positions for the dual-arm dexterity, denoted as $\hat{A}^{body}_t \in \mathbb{R}^{T \times 26}$.
\end{enumerate}

\noindent Formally, given a diffusion timestep $\tau$ and noisy action chunk $A_t^\tau$, the network $V_\theta$ approximates the noise $\epsilon$ by minimizing the decoupled denoising loss:
%
%
\begin{align}
\mathcal{L}_{diff} = \mathbb{E}_{\tau, \epsilon} \big[
    &\lambda_1 \left\| V_\theta^{cam}(\dots) - \epsilon^{cam} \right\|^2 \notag\\
    &+ \lambda_2 \left\| V_\theta^{body}(\dots) - \epsilon^{body} \right\|^2
\big]
\end{align}

\noindent where $V_\theta(\phi, A_t^\tau, q_t)$ conditions on the fused embedding $\phi$. This bifurcation enables the agent to learn high-frequency visual servoing (head) independently from complex manipulation planning (body).

\subsubsection{Camera Adapter with Eagle-2 and LoRA}

To imbue the agent with high-level semantic understanding for active search, we utilize \textbf{Eagle-2} as our Vision-Language Model (VLM) backbone. Eagle-2 integrates a SigLIP-2 image encoder with a SmolLM2 LLM, providing strong priors for grounding text instructions to visual observations.
We sought to preserve the original VLM’s high-level semantic information while efficiently learning an alignment between high-level semantics and low-level camera movements.Since full fine-tuning of the billion-parameter VLM is computationally prohibitive and risks catastrophic forgetting, we implement a \textbf{Camera Adapter} using Low-Rank Adaptation (LoRA). 
Referencing the parameter-efficient fine-tuning paradigm, we inject trainable rank decomposition matrices into the linear layers of the LLM's attention blocks while freezing the pre-trained weights $W_0$. For a linear projection $h = W_0 x$, the adapted forward pass is given by:
\begin{equation}
h = W_0 x + \frac{\alpha}{r} B A x
\end{equation}
where $B \in \mathbb{R}^{d \times r}$ and $A \in \mathbb{R}^{r \times k}$ are trainable matrices with rank $r \ll d$. 
                                                            This adapter allows SaPaVe to align the generic semantic features of Eagle-2 with the specific requirements of active robotic perception using less than 2\% of the trainable parameters, effectively bridging the gap between "internet-scale" knowledge and "robot-scale" control.

\subsubsection{Universal Spatial Knowledge Injection}

A core contribution of SaPaVe is the integration of explicit 3D geometric knowledge. We leverage the encoder from \textbf{MapAnything}, a foundation model capable of processing arbitrary combinations of visual and geometric modalities.

\paragraph{Multi-Modal Encoding Process.} 
As shown in the reference architecture, the MapAnything encoder processes $N$ input views. It handles two streams of data:
\begin{itemize}
    \item \textbf{Visual Stream:} RGB images are encoded using a frozen \textit{DINOv2} (ViT-L) backbone to extract high-level patch features $F_{rgb}$.
    \item \textbf{Geometric Stream:} Optional geometric priors (Camera Poses $\hat{P}$, Ray Directions $\hat{R}$, and Depths $\hat{D}$) are factorized and projected. Specifically, ray directions and normalized depths are processed via a shallow Convolutional Encoder with pixel unshuffle to match the specialized dimension of DINOv2 features. Global quantities (rotations, scales) are projected via a 4-layer MLP.
\end{itemize}
These features are fused via a sum-based aggregation:
\begin{equation}
F_{spatial} = \text{LayerNorm}(F_{rgb} + \text{Proj}(F_{geo}))
\end{equation}
This results in a dense spatial token representation $F_{spatial}$ that inherently encodes metric 3D structure (positions and occupancy) aligned with the visual patches.

\vspace{+1mm}
\noindent \textbf{Injection via Element-wise Addition and Cross-Attention.}
To inject this spatial awareness into the action policy, we employ a late-fusion strategy within the DiT blocks. 
Let $\phi_{vlm}$ be the semantic tokens output by the Camera Adapter (Eagle-2 + LoRA). We first align the dimension of the spatial tokens $F_{spatial}$ to match $\phi_{vlm}$ via a linear projection. The multi-modal conditioning context $\phi_{fused}$ is obtained by \textbf{Element-wise Addition}:
\begin{equation}
\phi_{fused} = \phi_{vlm} + \beta \cdot \text{Linear}(F_{spatial})
\end{equation}
where $\beta$ is a learnable scaling factor. Finally, this fused representation $\phi_{fused}$ serves as the \textit{Key} and \textit{Value} in the \textbf{Cross-Attention} layers of the Decoupled Action Heads (DiT), while the noised action tokens serve as the \textit{Query}. This mechanism ensures that every denoising step is guided by both the semantic intent (from VLM) and the precise 3D geometry (from MapAnything Encoders).

\subsection{Detailed Training Protocols}

To effectively learn the complex coordination between active perception (head movements) and dexterous manipulation (dual-arm control), we employ a curriculum-style \textbf{Two-Stage Training Strategy}. This approach prevents the policy from falling into local minima where the robot attempts to manipulate without first establishing a valid visual observation.

\subsubsection{Stage 1: Semantic Active Perception Alignment}
\textbf{Objective.} 
In the initial stage, our primary goal is to bridge the gap between the VLM's high-level semantic understanding and the low-level kinematic control of the active head. We aim to equip the model with a strong \textit{prior} for "where to look" based on language instructions, independent of the arm manipulation complexity.

\vspace{+1mm}
\noindent \textbf{Data Configuration.} 
We use proposed ActiveViewPose-200K dataset. Unlike standard manipulation datasets collected from fixed, near-optimal viewpoints, this dataset comprises 200k tuples of $(Image, Instruction, \text{CameraAction})$. It specifically targets the transition from suboptimal or out-of-view initial states to optimal task-oriented viewpoints.

\vspace{+1mm}
\noindent \textbf{Optimization Strategy.} 
During this stage, we freeze the \textit{Universal Spatial Encoder} (MapAnything) and the main \textit{Manipulation Action Decoder}. We strictly enforce updates on:
\begin{enumerate}
    \item The \textbf{Camera Adapter} (LoRA layers on Eagle-2), allowing the VLM to adapt to active viewing instructions.
    \item The \textbf{Camera Action Decoder} (the 2-DoF MLP head), learning to translate semantic features into pitch and yaw velocities.
\end{enumerate}

\vspace{+1mm}
\noindent \textbf{Loss Function.} 
The training objective minimizes the Mean Squared Error (MSE) between the predicted camera movement $\hat{A}_{\text{head}}$ and the ground-truth trajectory $A^*_{\text{head}}$:
\begin{equation}
\mathcal{L}_{\text{Stage1}} = \mathcal{L}_{\text{MSE}}(\hat{A}_{\text{head}}, A^*_{\text{head}}) = \frac{1}{T} \sum_{t=1}^{T} || \hat{A}_{\text{head}}^t - A_{\text{head}}^{*t} ||^2
\end{equation}
By the end of Stage 1, the model develops a robust capability to perform semantic active search, effectively becoming a "embodied cameraman" that can locate target objects specified by natural language.

\subsubsection{Stage 2: Active Manipulation Fine-tuning}
\textbf{Objective.} 
Building upon the semantic active perception prior established in Stage 1, the second stage introduces the full complexity of dexterous manipulation. This stage aims to achieve \textit{Active-View Execution}, where the robot coordinates its gaze to support hand-object interaction.

\vspace{+1mm}
\noindent \textbf{Data Configuration.} 
To prevent catastrophic forgetting of the active perception skills while learning manipulation, we employ a \textbf{Data Mixture Strategy}. The training batch consists of samples from both the \textit{ActiveViewPose-200K} dataset (to maintain look-at capability) and the \textit{Active Manipulation Robot Data} (generated via DexFlyWheel, covering the 12 core tasks),as well as the real world collected data.

\vspace{+1mm}
\noindent \textbf{Optimization Strategy.} 
In this stage, we unlock the \textbf{Decoupled Action Heads}. We jointly train the entire action generation pipeline, including both the Camera Action Decoder and the Manipulation (Body) Action Decoder.

\vspace{+1mm}
\noindent \textbf{Loss Function.} 
The objective is a weighted sum of the head movement loss and the body manipulation loss. We define the total loss as:
\begin{equation}
\mathcal{L}_{\text{Stage2}} = \lambda_{\text{head}} \mathcal{L}_{\text{head}} + \lambda_{\text{other}} \mathcal{L}_{\text{other}}
\end{equation}
where:
\begin{itemize}
    \item $\mathcal{L}_{\text{head}} = || \hat{A}_{\text{head}} - A^*_{\text{head}} ||^2$ ensures the camera continues to track relevant objects.
    \item $\mathcal{L}_{\text{other}} = || \hat{A}_{\text{body}} - A^*_{\text{body}} ||^2$ supervises the 26-DoF arm and hand joints.
    \item $\lambda_{\text{head}}$ and $\lambda_{\text{other}}$ are balancing coefficients. In practice, we set $\lambda_{\text{head}}=1.0$ and $\lambda_{\text{other}}=10.0$ to prioritize the higher-dimensional manipulation space while maintaining visual stability.
\end{itemize}
This two-stage approach ensures a "bottom-up" acquisition of skills: first learning to perceive, and then learning to act upon that perception.

\section{Additional Demonstrations}
\label{suppsec: more demonstrations}

In this section, we provide a comprehensive qualitative analysis of {\ourmodel} through extensive visualizations across both simulation and real-world environments.

\subsection{Simulation Demonstrations}
\begin{figure*}[t]
\centering
\includegraphics[width=\linewidth]{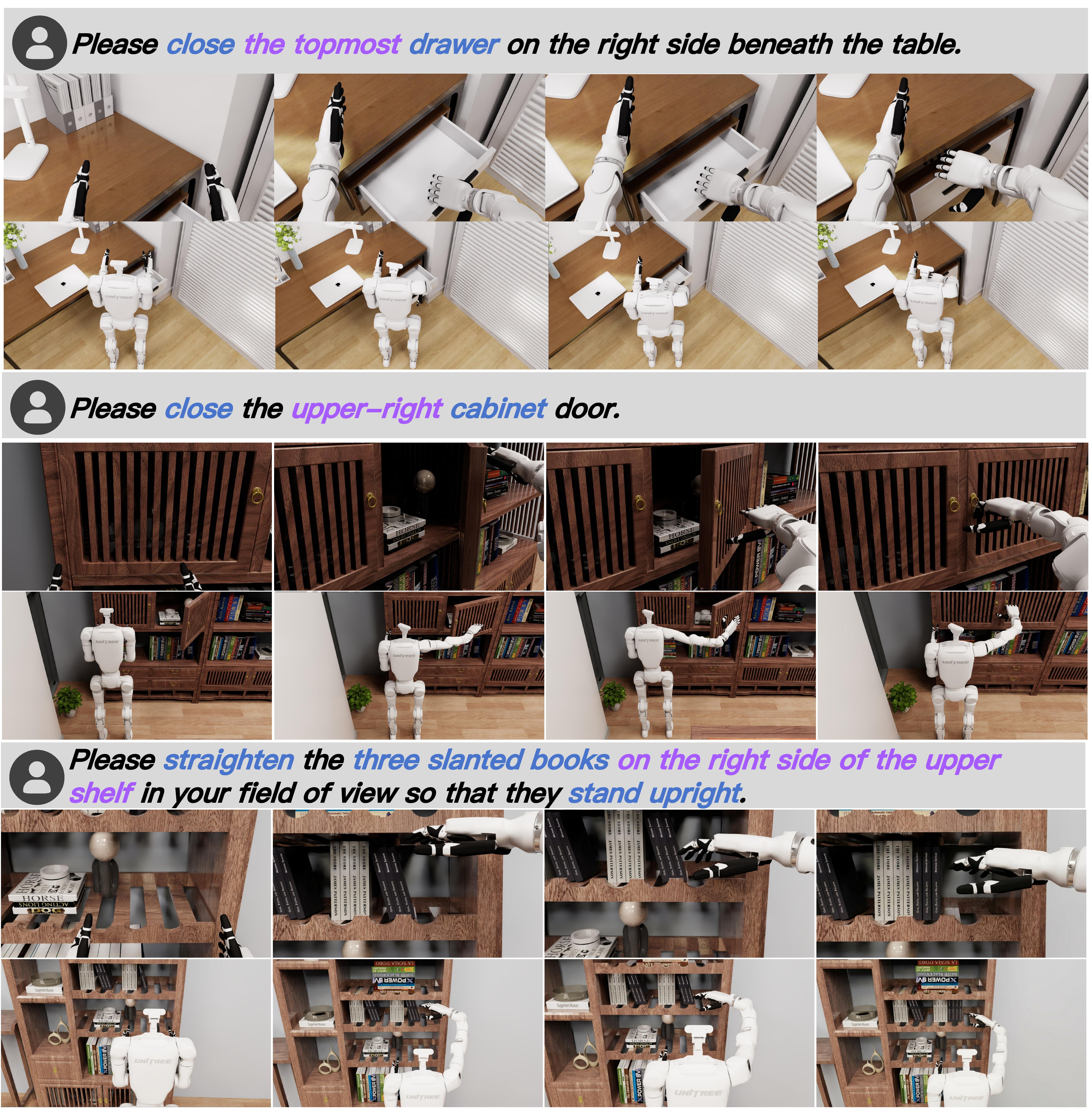}
\vspace{-6mm}
   \caption{This image illustrates the active manipulate task performed by the robot in a simulation scenario.
   }
\label{fig: supp_simbenchmark_2}
\vspace{-5mm}
\end{figure*}

\begin{figure*}[t]
\centering
\includegraphics[width=\linewidth]{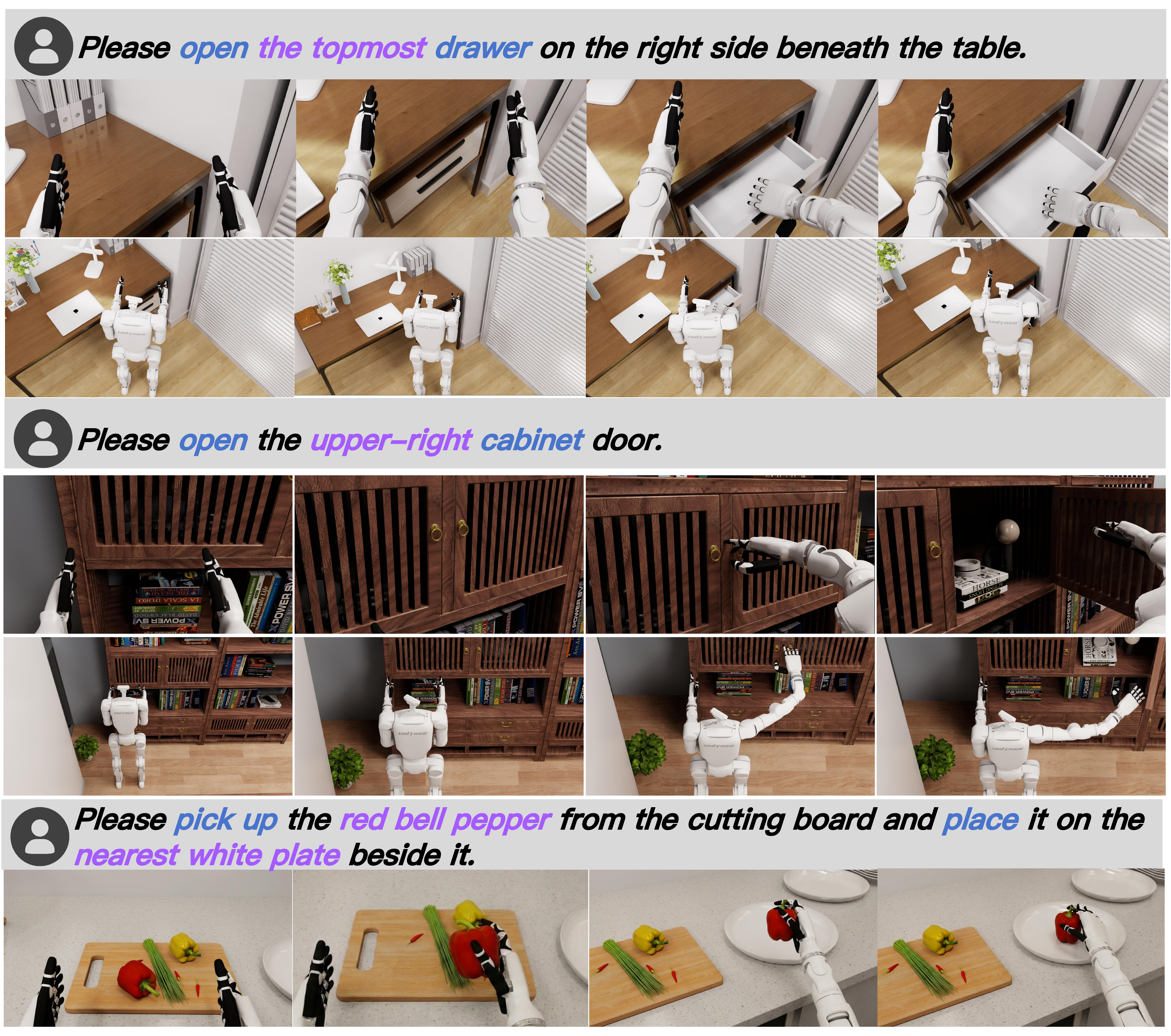}
\vspace{-6mm}
   \caption{This image illustrates the active manipulate task performed by the robot in a simulation scenario.
   }
\label{fig: supp_simbenchmark_1}
\vspace{-5mm}
\end{figure*}

We present time-lapse sequences of the 12 active manipulation tasks in Fig.~\ref{fig: supp_simbenchmark_2}~\ref{fig: supp_simbenchmark_1}. These visualizations highlight the distinct behavior of our model compared to passive baselines:
\begin{itemize}
    \item \textbf{Active View Finding:} In \textsc{Out-of-View} initialization settings, the agent is shown rapidly rotating its camera to scan the environment based on semantic instructions (e.g., looking down for "bottom cabinet") before minimizing arm movement.
    \item \textbf{Handling Occlusion:} In the \textsc{FetchFrom} tasks, the visualizations depict the agent actively maneuvering its head to peer inside opened drawers or cabinets to locate objects effectively before attempting a grasp.
\end{itemize}
Please refer to the supplementary video/website for full-length trajectory playbacks, which demonstrate the temporal smoothness and stability of the decoupled head-body coordination.

\subsection{Real-World Robot Deployment}
\begin{figure*}[t]
\centering
\includegraphics[width=\linewidth]{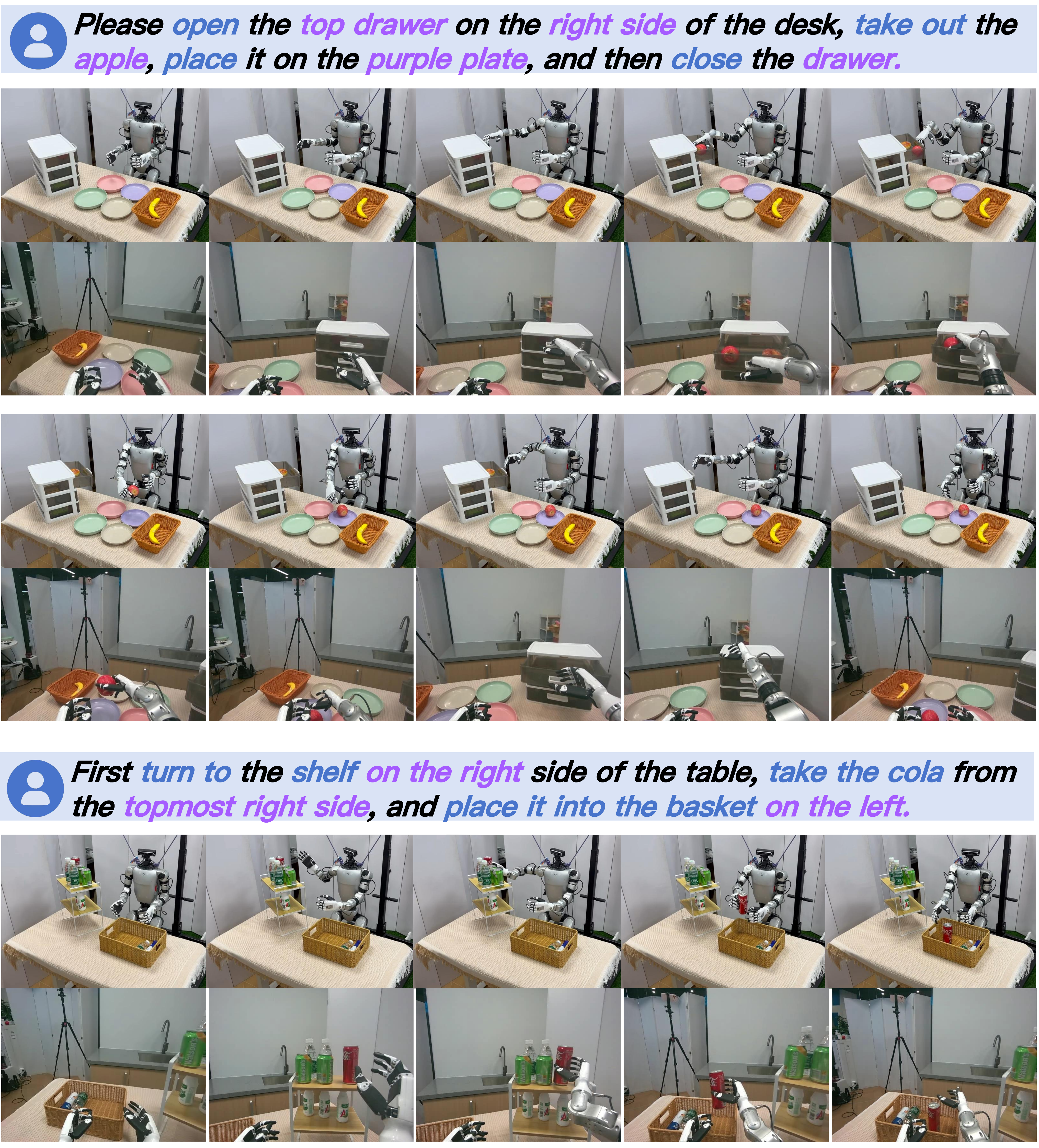}
\vspace{-6mm}
   \caption{This image illustrates the active manipulate task performed by the robot in a real-world scenario.
   }
\label{fig: supp_real_world_1}
\vspace{-5mm}
\end{figure*}

\begin{figure*}[t]
\centering
\includegraphics[width=\linewidth]{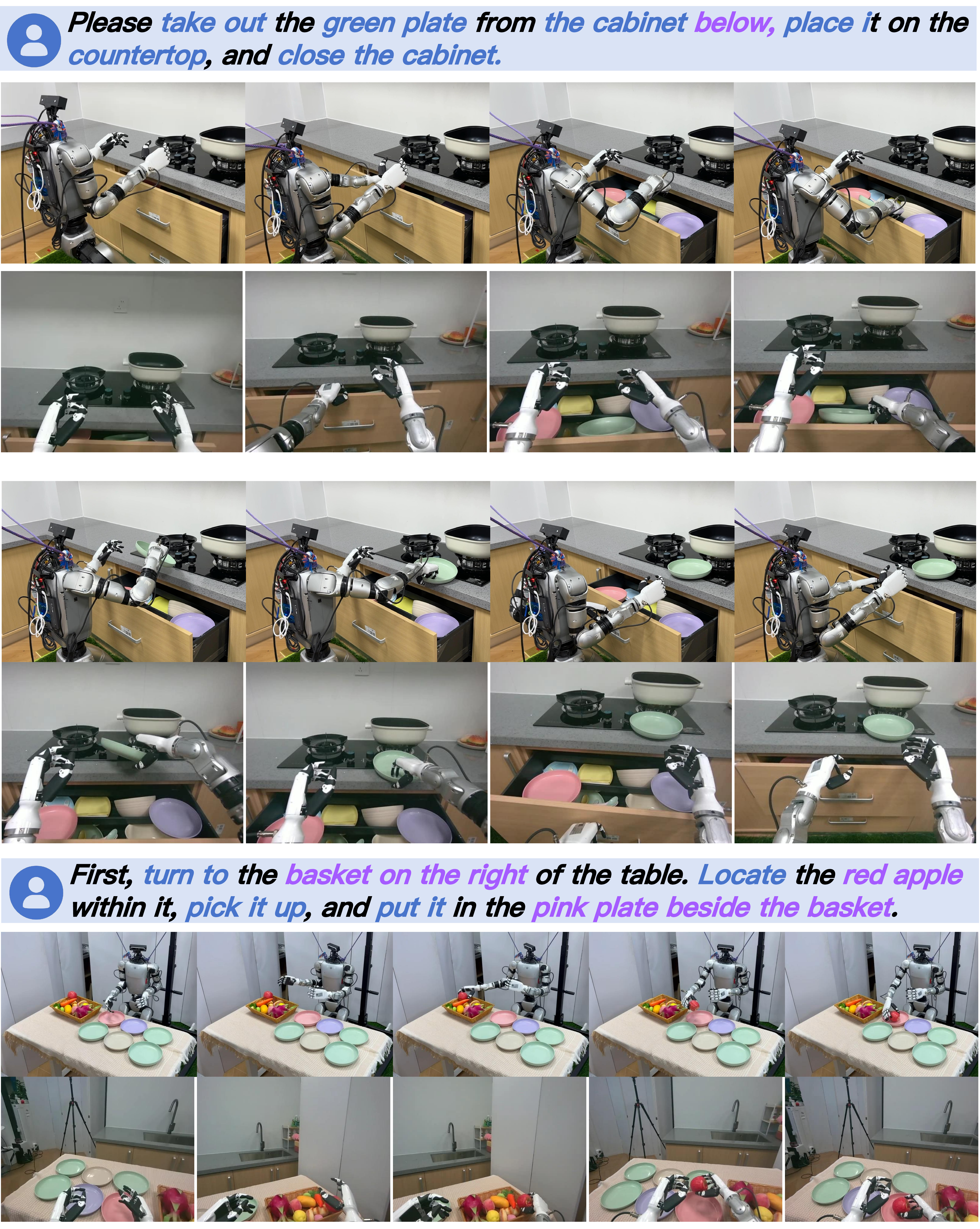}
\vspace{-6mm}
   \caption{This image illustrates the active manipulate task performed by the robot in a real-world scenario.
   }
\label{fig: supp_real_world_2}
\vspace{-5mm}
\end{figure*}

We further illustrate the robustness of {\ourmodel} on the Unitree G1 humanoid platform. Figure~\ref{fig: supp_real_world_1} ~\ref{fig: supp_real_world_2}showcases execution traces for the \textsc{Out-of-View Pick-and-Place} and \textsc{Occluded Articulated Manipulation} tasks.
Crucially, we demonstrate the model's ability to recover from disturbances. In scenarios where the target object is manually moved out of the current frame during execution, {\ourmodel} successfully re-initiates an active search to re-acquire the target, validating the closed-loop nature of our active perception pipeline.

\section{Limitations and Future Work}
\label{suppsec: limitation}

While {\ourmodel} demonstrates promising capabilities in active semantic manipulation, we acknowledge certain limitations in our current setup that point towards exciting directions for future research.

\paragraph{Workspace Constraints under Static Base.}
The primary limitation of the current implementation is that the robot operates with a fixed base. Although the active head significantly expands the effective \textit{perceptual} field of view, the \textit{manipulation} workspace remains bounded by the stationary reach of the robot's arms. 
Consequently, if an object is located or moved beyond the maximum arm span (even if actively perceived and located by the head), the agent cannot complete the task. The current active manipulation is thus confined to "local" exploration rather than "global" search.

\paragraph{Towards Mobile Active Manipulation.}
To address the aforementioned constraint, our future work aims to extend {\ourmodel} to Mobile Manipulation settings. 
We envision a holistic framework where active perception controls not only the head (pitch/yaw) but also the mobile base (navigation). This would enable the robot to:
\begin{enumerate}
    \item \textbf{Navigate to Look:} Move to different rooms or distinct areas of a room to locate objects described by high-level instructions (e.g., "Find the apple in the kitchen").
    \item \textbf{Approach to Manipulate:} Actively position its base to maximize manipulability scores for grasping objects in confined spaces (e.g., inside deep cabinets).
\end{enumerate}
Integrating whole-body control with VLM-guided active semantic search will be a critical step towards truly general-purpose embodied agents.

\newpage

\end{document}